\documentclass[a4paper,fleqn,numbers]{cas-dc}

\usepackage[numbers,sort&compress]{natbib}
\usepackage{amssymb}
\usepackage{amsmath}
\usepackage{ragged2e}
\usepackage{booktabs, multirow, xltabular, array, caption, xcolor}

\usepackage{array}
\usepackage{tabularx}
\usepackage{float}
\usepackage{stfloats}
\usepackage{subcaption}
\def\tsc#1{\csdef{#1}{\textsc{\lowercase{#1}}\xspace}}
\tsc{WGM}
\tsc{QE}
\begin{document}
\let\WriteBookmarks\relax
\def\floatpagepagefraction{1}
\def\textpagefraction{.001}

\shorttitle{}    

\shortauthors{Chi Liu, Yuzhuo Zhou, Sheng Shen, et al.}  

\title [mode = title]{Fundus Image-based Glaucoma Screening via Retinal Knowledge-Oriented Dynamic Multi-Level Feature Integration}  

% % Title footnote mark
% % eg: \tnotemark[1]

% % Title footnote 1.
% % eg: \tnotetext[1]{Title footnote text}

\author[cityu]{Chi Liu}%[]
% \cormark[1]
% Footnote of the second author
\fnmark[1]
% Email id of the second author
% \ead{chiliu@cityu.edu.mo}


\credit{Funding acquisition, Supervision, Conceptualization, Methodology, Writing - Review \& Editing}

\author[zoc,mzu]{Yuzhuo Zhou}%[<options>]
\fnmark[1]
\credit{Methodology, Validation, Investigation, Data Curation, Writing - Original Draft, Visualization}
\fntext[fn1]{These authors contributed equally.}

\author[tor]{Sheng Shen}%[]
\credit{Supervision, Methodology, Writing - Review \& Editing}

\author[monash]{Zongyuan Ge}%[]
\credit{Supervision, Conceptualization}

\author[cityu]{Fengshi Jing}%[]
\credit{Writing - Review \& Editing}

\author[zoc]{Shiran Zhang}%[]
\credit{Methodology}

\author[gzfp]{Yu Jiang}%[]
\credit{Writing - Review \& Editing}

\author[xiangya]{Anli Wang}%[]
\credit{Conceptualization, Methodology}

\author[cityu]{Wenjian Liu}%[]
\credit{Writing - Review \& Editing}

\author[xiangya]{Feilong Yang}%[]
\credit{Conceptualization, Methodology}

\author[cityu]{Tianqing Zhu}%[]
\credit{Conceptualization}

\author[zoc]{Xiaotong Han}%[]
\cormark[1]
% Footnote of the second author
% Email id of the second author
\ead{hanxiaotong2@gzzoc.com}
\credit{Supervision, Conceptualization, Writing - Review \& Editing}


\cortext[1]{Xiaotong Han is the corresponding author}



\affiliation[cityu]{organization={Faculty of Data Science, City University of Macau},
            addressline={Avenida Padre Tomás Pereira},
            city={Taipa},
            postcode={999078},
            state={Macao},
            country={China}}
            
\affiliation[zoc]{organization={State Key Laboratory of Ophthalmology, Zhongshan Ophthalmic Center, Sun Yat-sen University, Guangdong Provincial Key Laboratory of Ophthalmology and Vision Science, Guangdong Provincial Clinical Research Center for Ocular Diseases},
            city={Guangzhou},
            postcode={510000},
            state={Guangdong},
            country={China}}
            
\affiliation[mzu]{organization={Hainan International College, Minzu University of China},
            addressline={Li'an International Education Innovation Pilot Zone},
            city={Lingshui},
            postcode={572423},
            state={Hainan},
            country={China}}
            
\affiliation[tor]{organization={School of Information Technology, Faculty of Business and Hospitality, Torrens University Australia},
            addressline={196 Flinders St},
            % city={Taipa},
            postcode={3000},
            state={VIC},
            country={Australia}}

\affiliation[monash]{organization={AIM for Health Lab, Faculty of IT, Monash University, Australia},
            addressline={25 Exhibition Walk},
            city={Clayton},
            postcode={3800},
            state={VIC},
            country={Australia}}
            
\affiliation[gzfp]{organization={Department of Ophthalmology, Guangzhou First People’s Hospital, the Second Affiliated Hospital of South China University of Technology},
            addressline={No. 1, Panfu Road, Yuexiu District},
            city={Guangzhou},
            postcode={510000},
            state={Guangdong},
            country={China}}

\affiliation[xiangya]{organization={The Third Xiangya Hospital of Central South University},
            addressline={138 Tongzipo Road},
            city={Changsha},
            postcode={410013},
            state={Hunan},
            country={China}}

\begin{abstract}
While deep learning has advanced automated glaucoma screening via color fundus photography, existing purely data-driven models often overfit to confounding imaging artifacts and struggle to capture unpredictable pathological cues located beyond predefined anatomical boundaries. To address these limitations, we propose a retinal knowledge-oriented framework that synergizes dynamic multi-scale feature learning with domain-specific anatomical priors. The proposed architecture adopts a tri-branch structure to jointly model the global retinal context, the structural characteristics of the optic cup and disc, and dynamically cropped pathological regions. Specifically, we devise a Dynamic Window Mechanism (DWM) that adaptively discovers diagnostically informative patches via image-level supervision. Furthermore, we introduce a Knowledge-Enhanced Convolutional Block Attention Module (KE-CBAM) that explicitly incorporates retinal priors from a pre-trained foundation model RETFound to guide spatial attention, preventing the network from assigning spurious weights to irrelevant background noise. Extensive evaluations on the large-scale AIROGS dataset demonstrate that our method achieves a state-of-the-art AUC of 98.5\% and an accuracy of 94.6\%. More importantly, the integration of anatomical priors effectively mitigates the inherent class imbalance, significantly improving the detection of referable glaucoma cases. Additional validations on the SMDG-19 benchmark further confirm its superior cross-domain generalization, indicating that our contribution provides a robust, interpretable, and scalable solution for real-world clinical glaucoma diagnosis. Our code is available at \url{https://github.com/magiczhuo/Glaucoma-Detection}.

\end{abstract}

%\nocite{*}

% Keywords

\maketitle

\section{Introduction}
\label{sec:introduction}
Glaucoma represents the second leading etiology of irreversible blindness globally, posing a substantial threat to public health. The World Health Organization estimates that approximately 80 million individuals are currently affected by this condition~\cite{WHO2023}. As glaucoma is frequently asymptomatic during its nascent stages, patients are often diagnosed only after irreversible optic nerve damage has manifested, underscoring the criticality of early detection and intervention. In addition, epidemiological data indicate that the global prevalence rose from 60.5 million in 2010 to 79.6 million by 2020~\cite{Quigley2006}. The early screening needs and the expanding patient population necessitates the development of high-throughput and reliable diagnostic tools. Recently, artificial intelligence (AI) has shown great promise in medical image analysis \cite{yu2026quantum, yu2026alzheimer}, particularly in boosting the efficiency and accuracy of glaucoma screening based on color fundus photography (CFP), a non-invasive, cost-effective, and highly-accessible tool that enables large-scale screening \cite{wang2024economic, Huang2025Adecadeprogress}. With the trend, the robustness and generalization capability of CFP-based automated glaucoma screening across heterogeneous clinical datasets has emerged an important research direction \cite{yan2024prompt}.

Conventional human-based glaucoma screening relies on qualitative measurement of fundus images, which is human-subjective and heavily influenced by specialized training, clinical experience, and interpretative bias. Standard clinical indicators include horizontal and vertical cup-to-disc ratio (CDR) and neuro-retinal rim width ~\cite{tielsch1988intraobserver}. Quantifying these parameters requires nuanced visual judgment of cup morphology and rim configuration, both exhibiting significant inter-patient variability and inter-observer disagreement \cite{tielsch1988intraobserver}. The variability is further exacerbated by pathological pattern diversity and image quality fluctuation, compromising reliability of large-scale screenings, highlighting the necessity for objective, scalable automation. 

AI-driven approaches offer a promising avenue to mitigate human subjectivity and improve diagnostic consistency. Existing deep learning systems have dynamically automated glaucoma screening by classifying patients into "referable" or "non-referable" categories based on early pathological indicators such as subtle splinter hemorrhages around the optic disc (OD) and retinal nerve fiber layer (RNFL) defects at superotemporal or inferotemporal margins. However, the spatial randomness and ill-defined boundaries of these lesions pose significant generalization challenges~\cite{r13,r10}. Existing pattern recognition models that explored invariant mathematical descriptors and deep global feature representations ~\cite{el20213d,el2024automatic,el2021novel,sahmoudi2024new,r9} cannot be applied to CFP-based glaucoma screening directly, since they potentially neglect fine-grained structures of the optic cup (OC) and disc while introducing task-irrelevant background regions that heighten sensitivity to imaging noise. More sophisticated methods, such as attention modules \cite{woo2018cbam} and multi-branch models \cite{liu2022label,zhou2025enhancing}, rely on feature representations of closed clean training sets, struggling to distinguish reliable pathological patterns from confounding imaging artifacts such as reflections, over-exposure, and uneven illumination. Without prior knowledge of ocular anatomy, these generic modules are prone to overfitting spurious correlations, mistakenly assigning high activation weights to clinically irrelevant background noise.

\begin{figure}[]
    \centering
    \includegraphics[width=\linewidth]{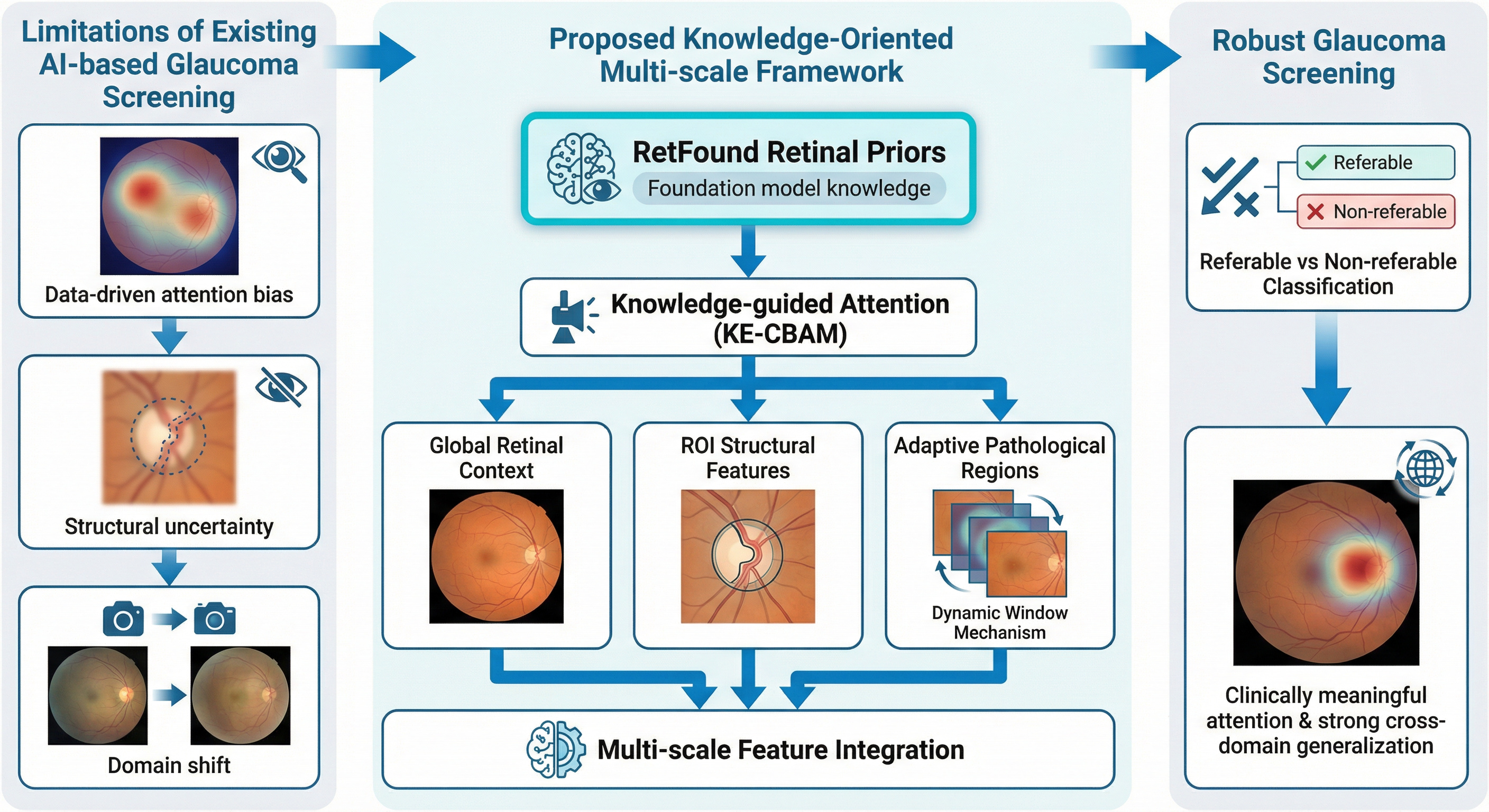}
    \caption{Conceptual illustration of the proposed retinal knowledge-oriented glaucoma screening framework. Existing deep learning models rely primarily on purely data-driven attention mechanisms, which suffer from limited anatomical awareness, structural uncertainty in optic disc–cup regions, and poor generalization across heterogeneous imaging devices. To address these challenges, the proposed framework integrates retinal priors extracted from the RETFound foundation model to guide attention learning. Knowledge-guided attention enables the model to jointly capture global retinal context, structural features of the optic disc–cup complex, and dynamically localized pathological regions. By combining retinal knowledge injection with multi-scale feature integration, the proposed framework achieves robust and interpretable glaucoma screening across diverse clinical datasets.}
    \label{fig:concept}
\end{figure}

To overcome this limitation, we incorporate RETFound as a domain-specific anatomical prior. Pre-trained on millions of unlabeled fundus images via self-supervised learning, RETFound encapsulates generalized, robust representation of retinal structures, guiding the network to constrain spatial focus within pathologically relevant regions. As illustrated in Figure~\ref{fig:concept}, the proposed framework jointly captures global retinal context, anatomical structures of the optic disc–cup complex, and dynamically localized pathological regions through a tri-branch architecture. The global branch encodes holistic retinal context from the full fundus image, while the ROI branch focuses on structural characteristics of the optic disc–cup complex. A Dynamic Window Mechanism (DWM) adaptively localizes diagnostically informative regions beyond predefined anatomical boundaries, capturing subtle pathological cues across heterogeneous retinal regions. We introduce a Knowledge-Enhanced Convolutional Block Attention Module (KE-CBAM), incorporating retinal structural priors from RETFound into the conventional channel–spatial attention framework. By integrating large-scale ophthalmic knowledge representations with task-specific visual features, KE-CBAM emphasizes clinically meaningful anatomical structures while suppressing irrelevant background patterns.

Through the synergistic integration of multi-scale retinal representations, adaptive pathological localization, and knowledge-guided attention learning, the proposed framework provides a robust and interpretable solution for automated glaucoma screening across heterogeneous clinical datasets. The main contributions of this work are summarized as follows:

\begin{itemize}

\item We propose a retinal knowledge-oriented glaucoma screening framework that integrates multi-scale retinal representations with domain-specific anatomical priors. The proposed tri-branch architecture jointly models global retinal context, OC/OD structures, and dynamically localized pathological regions.

\item We introduce a DWM that adaptively selects diagnostically informative regions beyond predefined anatomical regions, enabling the model to capture subtle pathological cues distributed across heterogeneous retinal images.

\item We propose KE-CBAM, which incorporates retinal anatomical priors extracted from the RETFound foundation model into a conventional channel-spatial attention mechanism, thereby enabling domain-aware attention learning.

\item Extensive experiments on the large-scale AIROGS dataset and multiple datasets from the SMDG-19 benchmark demonstrate that the proposed framework achieves superior diagnostic performance, strong cross-domain generalization, and consistent robustness under diverse imaging conditions, compared with state-of-the-art attention-based glaucoma screening methods.

\end{itemize}

\section{Related Work}

\subsection{Clinical Approaches for Glaucoma Screening}
Clinical approaches for glaucoma grading rely heavily on measuring image biomarkers associated with glaucomatous structural and functional impairment.  Jonas et~al.~\cite{Jonas1988} conducted a seminal morphometric study that established normative ranges for the OD, OC, and neuroretinal rim, providing essential quantitative baselines for horizontal and vertical cup–disc ratios (CDR). Garway–Heath et~al.~\cite{GarwayHeath1998} further elucidated the correlation between glaucomatous visual function loss and structural alterations in the optic nerve head (ONH) and peripapillary retinal nerve fiber layer (RNFL), and developed a precise anatomical mapping between visual field test locations and specific ONH sectors. More recently, Caiado et al.~\cite{Caiado2025} utilized automated colorimetric analysis via the Laguna ONhE software to quantify ONH hemoglobin (ONH Hb) levels, evaluating their relationship with structural and functional parameters. 

Despite the identification of these structural and functional indicators, clinical measurements are labor-intensive, and remain susceptible to inter-individual anatomical variability, vascular distribution differences, and subjective interpretative bias, resulting in only moderate inter-observer agreement~\cite{tielsch1988intraobserver}. This limitation has catalyzed the development of semi-automated diagnostic methods. For instance, Joshi et~al.~\cite{Joshi2011} combined local image features with vessel-bend detection for cup boundary extraction. Cheng et~al.~\cite{Cheng2013} proposed a superpixel-based framework for joint segmentation that incorporates statistical features and a self-assessment reliability score. However, these methods often rely on handcrafted features or fixed imaging priors, which constrains their robustness across different devices. 

\subsection{Deep Representation Learning for Glaucoma Screening}
Building on clinical insights into the relationship between glaucoma and the optic cup/disc (OC/OD), several studies apply deep learning to learn local representations around these regions. Li et al.~\cite{r25} proposed an end-to-end region-based CNN with dedicated proposal networks for OC and OD detection. Xu et al.~\cite{xu2012efficient} developed an interest mechanism based on spatial relevance cues to automatically localize OC and OD. Although these local-feature methods effectively characterize highly variable boundary regions, their limited receptive fields restrict global context, resulting in representations that are sensitive to regional imaging variations around OC/OD.

In contrast, some approaches process the entire fundus image as input~\cite{r9}. For example, Huang et al.~\cite{r9} introduced a dynamic local-to-global learning module using deformable convolution to adaptively focus on discriminative regions in low-resolution images. While capturing broader context, they may overlook fine-grained structures near OC/OD and introduce irrelevant background regions, increasing sensitivity to imaging noise such as overexposure or shadows.

Moreover, both local and global representation learning approaches lack a smooth and dynamic integration between the two. Their purely data-driven nature also tends to overlook valuable clinical domain knowledge.

\subsection{Retinal Priors from Foundation Models}
In recent years, foundation models have introduced new paradigms for retinal image analysis.
For example, Jalili et al.~\cite{r8} utilized GPT-4V glaucoma diagnosis via prompt-guided clinical rules. This method relies on manual optic disc cropping, thereby discarding the global retinal context—and suffer from inconsistent inferences.
Zhou et al.\cite{zhou2023foundation} proposed RETFound, the first ophthalmology-specific foundation model pretrained on 1.6 million unlabeled retinal images using self-supervised learning. Hou et al.\cite{hou2025comparison} compared general-purpose foundation models with RETFound across multiple ocular tasks and showed that RETFound provides stronger structural retinal priors and better generalization. Motivated by these findings, our study incorporates RETFound to provide external anatomical retinal priors as clinical knowledge guidance.

\begin{table*}[!t]
\centering
\caption{Comparison of representative existing methods and the proposed framework, highlighting the research gap in glaucoma screening.}
\label{tab:literature_comparison}
\renewcommand{\arraystretch}{1.15}
\setlength{\tabcolsep}{4pt}

% Single-column: scale to \linewidth (not \textwidth)
\resizebox{\textwidth}{!}{%
\begin{tabular}{@{}llcccl@{}}
\toprule
\textbf{Category} & \textbf{Reference / Method} & \textbf{Input Focus} & \textbf{Domain Prior} & \textbf{Attention Type} & \textbf{Key Limitations} \\
\midrule

\multirow{2}{*}{\begin{tabular}[c]{@{}l@{}}Clinical \&\\ Traditional\end{tabular}}
& Caiado et al.~\cite{Caiado2025} (Laguna ONhE) & Local (ONH) & Colorimetric & None & Sensitive to imaging variability; subjective interpretation. \\
& Joshi et al.~\cite{Joshi2011}, Cheng et al.~\cite{Cheng2013} & Local (OC/OD) & Handcrafted & None & Fixed handcrafted features; limited generalization. \\
\midrule

\multirow{2}{*}{\begin{tabular}[c]{@{}l@{}}Deep Learning\\ (Local-based)\end{tabular}}
& Li et al.~\cite{r25} (Region-based CNN) & Local (OC/OD) & Spatial & Disc-based & Loses global context; OD/OC variability sensitive. \\
& Xu et al.~\cite{xu2012efficient} & Local (OC/OD) & Structural & None & Fixed ROIs; misses peripheral cues. \\
\midrule

\multirow{1}{*}{\begin{tabular}[c]{@{}l@{}}DL (Global-based)\end{tabular}}
& Huang et al.~\cite{r9} (CNN-ViT \& Deformable) & Global & None & Generic (Multi-Head) & Prone to artifact-driven attention. \\
\midrule

\multirow{3}{*}{\begin{tabular}[c]{@{}l@{}}Foundation\\ Models\end{tabular}}
& Jalili et al.~\cite{r8} (GPT-4V) & Local & Prompt-based Clinical & Generic (Transformer) & Limited global context; inconsistent outputs. \\
& Zhou et al.~\cite{zhou2023foundation} (RETFound) & Global & Self-supervised & Generic (ViT) & Needs adaptation; weak local cue integration. \\
& Hou et al.~\cite{hou2025comparison} & Global & Foundation (Retinal \& General) & Generic (ViT) & Global fine-tune only; no adaptive localization. \\
\midrule

\textbf{Proposed} & \textbf{Tri-branch + DWM + KE-CBAM}
& \textbf{\begin{tabular}[c]{@{}c@{}}Hybrid\\ (Global+Local+Dynamic)\end{tabular}}
& \textbf{\begin{tabular}[c]{@{}c@{}}Anatomical\\ (RETFound)\end{tabular}}
& \textbf{\begin{tabular}[c]{@{}c@{}}Domain-oriented\\ (KE-CBAM)\end{tabular}}
& \textbf{\begin{tabular}[c]{@{}c@{}}Dynamic lesion localization with anatomical priors \\ to suppress spurious attention.\end{tabular}} \\
\bottomrule
\end{tabular}%
}
\end{table*}

\subsection{Research Gap}
To systematically summarize the aforementioned literature and clearly delineate the current research gaps, Table~\ref{tab:literature_comparison} provides a comprehensive comparison of representative glaucoma screening methods. As illustrated, while existing approaches have made significant advance, they typically suffer from at least one of the following limitations: (1) relying solely on either local or global representations without dynamic integration, (2) utilizing purely data-driven attention mechanisms may attribute to imaging artifacts, (3) lacking the guide of ophthalmologically anatomical priors to suppress spurious attention weights.

In the table of these identified gaps, it indicates that a robust glaucoma screening system requires a paradigm that simultaneously captures multi-scale spatial information and incorporates domain-specific clinical priors to overcome the spurious cues inherent in purely data-driven models. This exact gap directly motivates the design of our proposed retinal knowledge-oriented framework.

\section{Method}

\begin{figure*}
    \centering
    \includegraphics[width=\textwidth]{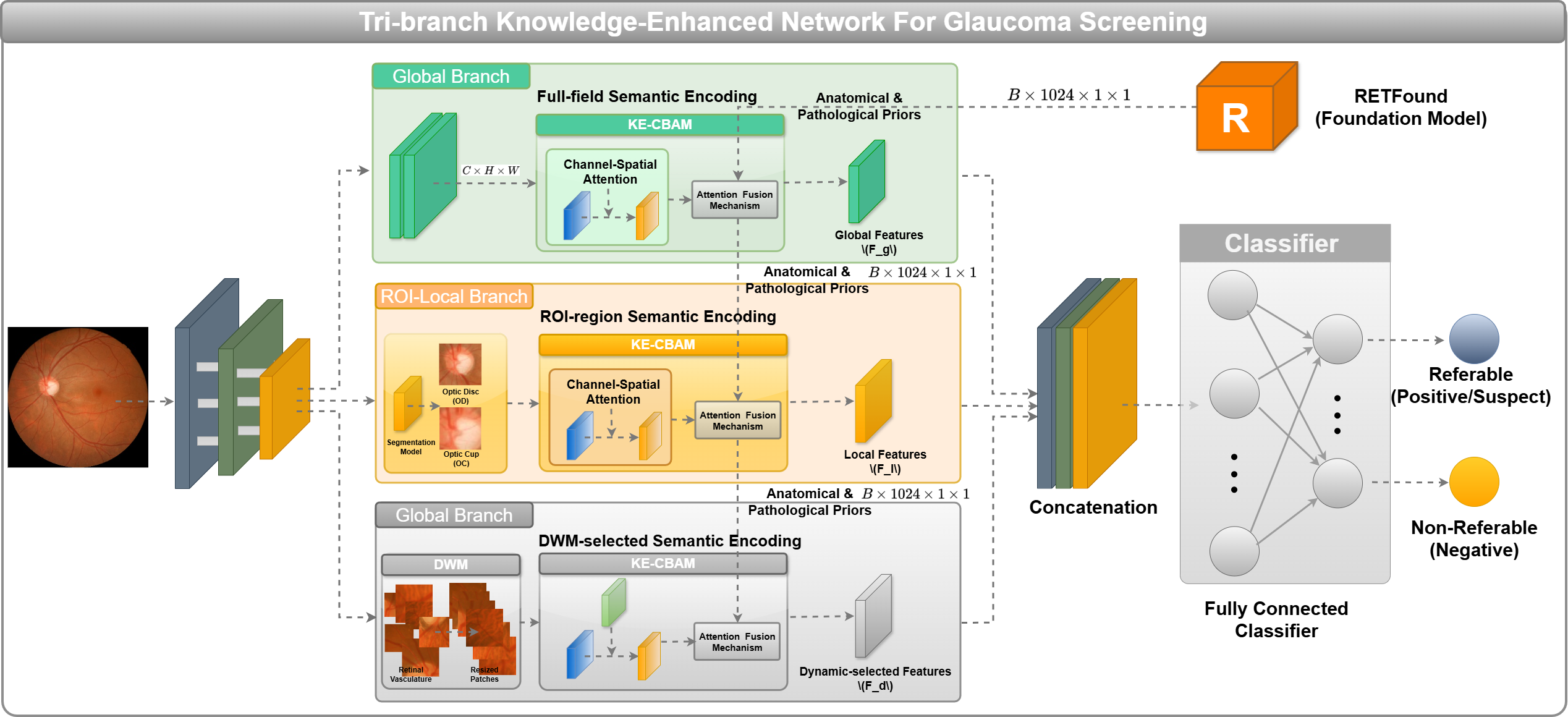}
    \caption{Overview of the proposed retinal knowledge-oriented glaucoma screening framework. 
The architecture adopts a tri-branch structure consisting of a global context branch, an ROI structural branch, and a dynamic window localization branch. 
Each branch employs a ResNet152 backbone enhanced with the proposed KE-CBAM, which integrates retinal anatomical priors extracted from the RETFound foundation model. 
The multi-scale feature representations are fused to produce the final glaucoma screening prediction.}        
    \label{fig:framework}
\end{figure*}

\subsection{Problem Formulation and Design Principles}

Automated glaucoma screening from color fundus photography requires the reliable identification of subtle structural abnormalities around the optic nerve head, particularly the optic disc (OD), optic cup (OC), and surrounding neuroretinal rim. However, the appearance of glaucomatous features varies substantially across patients, imaging devices, and acquisition conditions. These variations introduce significant challenges for deep learning systems trained on limited datasets, often leading to reduced cross-domain generalization and unstable feature representations.

Existing deep learning methods typically adopt either global image representations or localized region-based approaches. Global methods capture holistic retinal context but often include task-irrelevant background regions, which may introduce noise caused by illumination variation, shadows, or imaging artifacts. Conversely, local approaches concentrate on the optic disc region and its surrounding structures but may overlook pathological cues appearing outside predefined anatomical boundaries. In addition, most existing attention mechanisms are purely data-driven and lack explicit integration of ophthalmic domain knowledge, which limits their ability to focus on clinically meaningful structures.

To address these limitations, we propose a knowledge-oriented multi-scale framework guided by three design principles:

\begin{enumerate}

  \item \textbf{Clinical structural hierarchy modeling.}
  Glaucoma-related cues exist at multiple anatomical scales, from global retinal patterns to fine structural variations within the optic disc--cup complex. Feature extraction should simultaneously capture global context and localized anatomical structures.

  \item \textbf{Adaptive pathological localization.}
  Pathological regions may not coincide with predefined anatomical regions. A dynamic localization strategy is required to automatically identify diagnostically relevant image regions.

  \item \textbf{Knowledge-guided attention learning.}
  Generic attention modules lack domain awareness and may emphasize visually salient but clinically irrelevant regions. Incorporating retinal structural priors from large-scale ophthalmic datasets can guide attention toward clinically meaningful features.

\end{enumerate}

Based on these principles, we design a tri-branch architecture equipped with a knowledge-enhanced attention mechanism to integrate global contextual information, anatomical region features, and dynamically localized pathological cues. 

\subsection{Overview of the Proposed Framework}
The overall architecture of the proposed retinal knowledge-oriented glaucoma screening network is illustrated in Figure \ref{fig:framework}. Given an input fundus image $X \in \mathbb{R}^{3 \times H \times W}$, the network extracts multi-scale representations using three complementary branches 

\begin{itemize}
  \item a \textbf{global context branch}, encoding holistic retinal structures and overall fundus morphology for comprehensive full-image understanding;
  \item a \textbf{region-of-interest (ROI) branch}, focusing on the optic disc-cup complex, the primary anatomical region for glaucoma assessment, enabling fine-grained analysis of cup-to-disc ratio variations;
  \item a \textbf{dynamic window branch}, adaptively localizing high-response pathological regions to capture spatially distributed and subtle pathological cues that may be overlooked by fixed-region strategies.
\end{itemize}

All three branches share a ResNet-152 backbone for feature extraction. To improve feature discrimination and domain robustness, a \textbf{KE-CBAM} is embedded within each branch, incorporating retinal anatomical priors from RETFound to guide channel--spatial attention learning, directing focus toward clinically relevant regions while suppressing diagnostically irrelevant noise.

The feature embeddings from the three branches are fused to form a unified representation, which is fed to a fully connected classification head for glaucoma detection.

\subsection{Multi-scale Feature Representation}

\subsubsection{Global Context Branch}

The global branch processes the entire fundus image to capture holistic anatomical structures and contextual patterns that may indicate glaucomatous changes.

The input image $X \in \mathbb{R}^{3 \times H \times W}$ is passed through the ResNet-152 backbone to generate a feature map $F \in \mathbb{R}^{C \times H' \times W'}$, where $C$ denotes the channel dimension and $H', W'$ represent the spatial resolution after downsampling. To perform progressive feature transformation and spatial scale adjustment, the convolutional feature encoder is followed by four residual module with different structural configurations. Within each residual module, feature mapping is governed by specific convolutional kernels and strides. To obtain a compact representation, global average pooling is applied, followed by a fully connected layer $\mathrm{FC}(\cdot)$ that projects the result into the embedding space:
\begin{equation}
f_g = FC(AvgPool(F))
\end{equation}
where $f_g$ denotes the global feature embedding.

\subsubsection{ROI Structural Branch}

The ROI branch focuses on fine-grained structural details around the optic disc and optic cup, which are key anatomical indicators for glaucoma diagnosis. Compared with the global branch, the ROI branch preserves more spatial details to better capture morphological variations of the cup-to-disc ratio and neuroretinal rim.

The OD--OC region is first segmented using a pretrained segmentation model~\cite{Fu2018}. The cropped region $X_{roi} \in \mathbb{R}^{3 \times H_r \times W_r}$ is fed into the same backbone architecture to extract high-resolution structural features $F_1 \in \mathbb{R}^{C \times H' \times W'}$. The final ROI feature embedding is computed as
\begin{equation}
f_r = FC(AvgPool(F_1))
\end{equation}

\subsubsection{Dynamic Window Localization Branch}

\begin{figure}
    \centering
    \includegraphics[width=\linewidth]{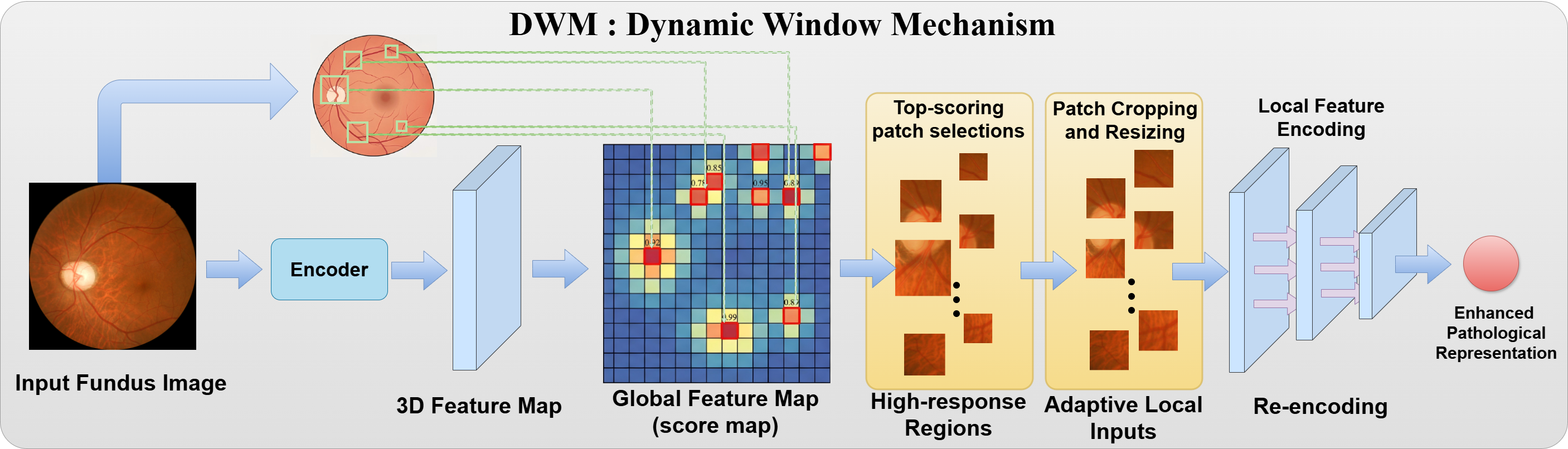}
    \caption{Illustration of the DWM. High-response regions are identified from the global feature map using response score estimation. The top-scoring patches are cropped and resized to form adaptive local inputs, enabling the model to capture subtle pathological cues beyond predefined anatomical regions.}
    \label{fig:dwm}
\end{figure}

Although the ROI branch captures anatomical structures around the optic disc, glaucomatous indicators, such as retinal nerve fiber layer defects or peripheral hemorrhages, may also appear in other diverse retinal regions. To address this issue, we introduce a \textbf{DWM}, illustrated in Figure~\ref{fig:dwm}, to autonomously localize high-response regions within the global feature map, thereby complementing the fixed ROI branch with image self-adaptive local evidence.

As indicated in the DWM pipeline, the input fundus image is first processed by the backbone encoder to produce a 3D global feature map. Given the global feature map $F \in \mathbb{R}^{C \times H' \times W'}$, average pooling is first applied to generate a 2D response score map $S \in \mathbb{R}^{H_s \times W_s}$ to evaluate the pathological relevance of different spatial locations. 

Specifically, for each scale $m \in \{1,2\}$, the feature map is smoothed by average pooling with kernel size $k_m$, and a score map is obtained by summing over the channel dimension:
\begin{equation}
S^{(m)} = \sum_{c=1}^{C} \mathrm{AvgPool}_{k_m}(F_c),
\end{equation}
where $k_1 = 3$ and $k_2 = 2$ in our implementation. This design encourages the model to identify spatial regions with consistently strong activation rather than relying on a single pixel-level peak. To further suppress redundant maxima, non-maximum suppression is applied on each score map, and the top responses are retained as candidate localization centers. In our implementation, each scale contributes three candidate windows, yielding six windows in total.

For a selected peak location $(i_{max}, j_{max})$ on the score map, the normalized location in the original image is computed by mapping the response coordinates back to the spatial extent of the input image that illustrated by the green dashed lines in Figure~\ref{fig:dwm}.
The spatial dimensions of the score map is defined as $H_s \times W_s$. The normalized center coordinates $(x_{center}, y_{center})$ of the target patch are calculated as follow: 

\begin{equation}
\left\{
\begin{aligned}
x_{center} &= \frac{2i_{max} + H - H_s + 1}{2H} \\
y_{center} &= \frac{2j_{max} + W - W_s + 1}{2W}
\end{aligned}
\right.
\end{equation}

The top-left and bottom-right boundaries of the local window are then given by
\begin{equation}
\begin{aligned}
x_{t} = x_{center} - \frac{h_p}{2}, \quad x_{b} = x_c + \frac{h_p}{2}, \\
y_{l} = y_{center} - \frac{w_p}{2}, \quad y_{r} = y_c + \frac{w_p}{2},
\end{aligned}
\end{equation}
where $(h_p,w_p)$ is the patch size associated with the selected scale. In our implementation, the two candidate patch sizes are $(224,224)$ and $(112,112)$, which correspond to the two pooling scales. Boundary correction is then applied so that the cropped window remains within the valid image range.

Each selected window $X_i^{patch}$ is cropped dynamically from the original image and resized to the network input resolution before being re-encoded by the backbone. The resulting local embeddings are denoted by $F_2^i$ $\{F_2^i\}_{i=1}^{p}$, where $p=6$ in our implementation. These local embeddings are fused with the global whole-image embedding and the ROI embedding through a lightweight multi-branch aggregation module:
\begin{equation}
f_d = FC \left( AvgPool \left( \sum_{i=1}^{p} F_2^i \right) \right)
\end{equation}
In the final classifier, the dynamic window branch is not supervised by lesion-level annotations. Instead, it is supervised by anatomical priors from the foundational model RETFound, and the localization behavior emerges from the end-to-end classification objective. Although early training may assign high responses to anatomically salient but non-pathological regions, such as the macula or specular reflections, these responses are progressively calibrated by the classification loss and the multi-branch fusion with the whole-image and ROI representations. In this way, the DWM branch learns to prioritize pathological regions that are effective for the final diagnostic decision.

The RETFound features utilized in this branch serve as static anatomical priors. For each input fundus image, a 1000-dimensional embedding is precomputed offline and cached to be reused throughout training and inference. During the forward pass, the network dynamically generates a fixed number $p=6$ of localized windows for each image. These cropped windows are subsequently stacked along the batch dimension, yielding a localized input tensor expanded to $(B \times p, 3, 299, 299)$, where $B$ denotes the original mini-batch size. 

To systematically align the offline global prior with these dynamically generated input tensors, the corresponding RETFound embeddings are expanded via a tensor interleaving operation. Specifically, the initial $(B, 1000)$ prior tensor is replicated along the batch dimension by a factor of $P$, with the final dimension of $(B \times P, 1000)$. This broadcast mechanism guarantees that all dynamic windows originating from the same parent image inherit the exact same prior, strictly matching the sequential order of the expanded window tensor. Consequently, the alignment between the offline prior and the dynamic localization process is established through sample-index broadcasting rather than spatial coordinate binding. This design preserves the spatial flexibility of the network to learn adaptive patch locations.

In our implementation, the DWM branch uses two pooling scales, three candidate windows per scale, and patch sizes of $224 \times 224$ and $112 \times 112$. The cropped windows are resized to $299 \times 299$ before being forwarded through the backbone. RETFound features are pre-extracted offline as 1000-dimensional vectors and are repeated across all windows of the same image during the forward pass. 
To facilitate qualitative interpretation, we further visualize the top-ranked windows selected by DWM in Section~\ref{Visualization}. 

\begin{figure}
    \centering
    \includegraphics[width=\linewidth]{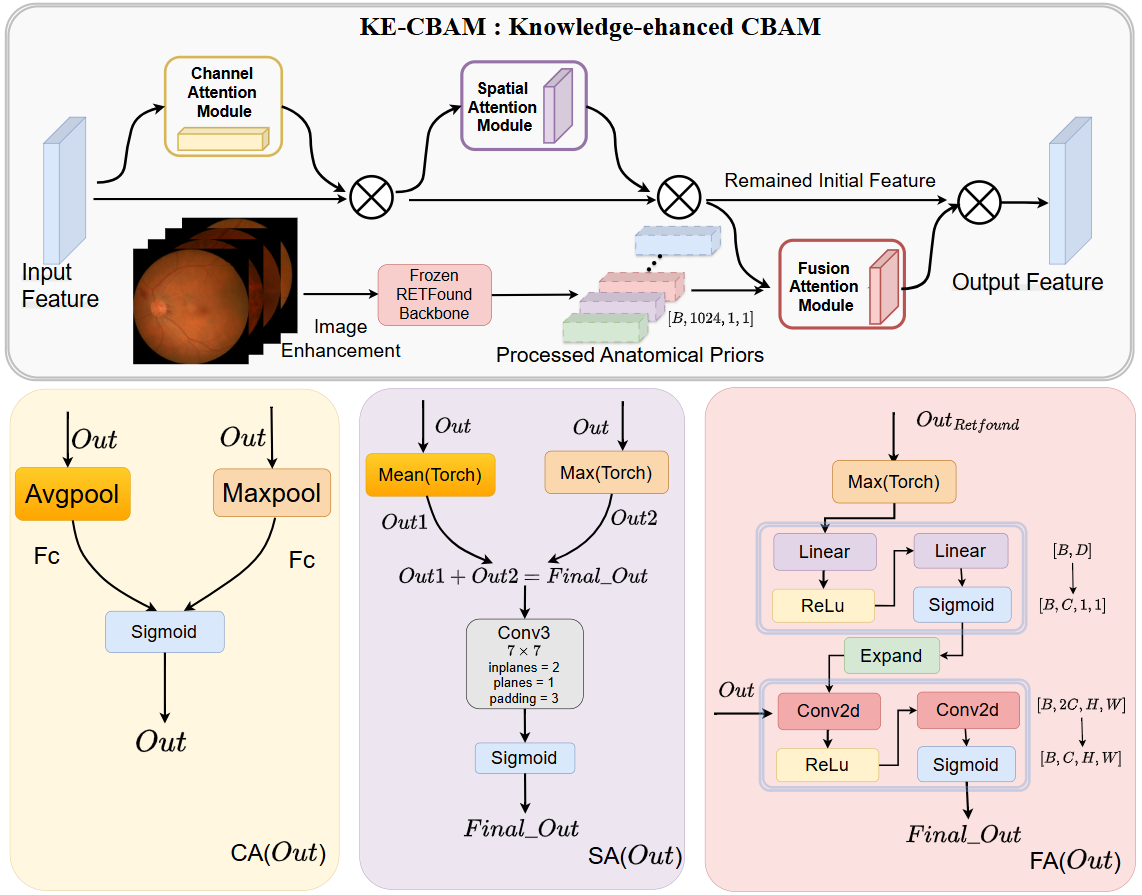}
    \caption{Workflow of the proposed KE-CBAM. 
Retinal anatomical priors extracted from the RETFound foundation model are projected into the backbone feature space and fused with CBAM-generated channel–spatial features through a cross-modal attention mechanism. 
This process enables the network to incorporate domain-specific retinal knowledge into the feature weighting process, improving attention localization toward clinically relevant structures.
}
    \label{fig:kecbam}
\end{figure}

\subsection{Knowledge-Enhanced Convolutional Block Attention Module}
To incorporate retinal domain priors into feature learning, we propose the \textbf{KE-CBAM}, which augments each bottleneck of a ResNet-152 backbone with residual attention guided by a pretrained retinal foundation model. In contrast to conventional CBAM, which performs only sequential channel and spatial refinement, KE-CBAM explicitly injects RETFound-based knowledge into the feature representation, thereby making the attention process more domain-aware and interpretable.

As illustrated in Figure~\ref{fig:kecbam}, each fundus image is first processed by an offline RETFound feature extraction pipeline. The image is resized to $224 \times 224$ and normalized before being passed through a pretrained RETFound encoder. During this extraction stage, probabilistic quality enhancement is applied, including CLAHE, contrast and brightness adjustment, sharpening, denoising, gamma correction, color enhancement, edge enhancement, and multi-scale fusion. 
The pipeline yields a 1000-dimensional retinal embedding, denoted as $F_{rf}$, which serves as a frozen domain prior rather than a trainable representation. Consequently, RETFound provides stable retinal semantics without introducing optimization instability into the KE-CBAM branch.

Let the feature map produced by the CBAM-refined backbone block be denoted as $F_{cbam} \in \mathbb{R}^{B \times C \times H \times W}$, where $B$ is the batch size, $C$ is the channel dimension, and $H$ and $W$ are the spatial dimensions. To align the RETFound embedding with the backbone representation, $F_{rf}$ is first projected by a two-layer bottleneck mapping with reduction ratio $r=16$:
$$
G = \sigma \left( W_2 \, \delta \left( W_1 F_{rf} \right) \right),
$$
where $\delta(\cdot)$ denotes the ReLU activation and $\sigma(\cdot)$ denotes the sigmoid function. The resulting vector $G$ is then reshaped and broadcast over the spatial dimensions to obtain $G_{exp} \in \mathbb{R}^{B \times C \times H \times W}$.

Next, the backbone feature map and the expanded prior are concatenated along the channel dimension and passed through a lightweight fusion branch composed of two $1 \times 1$ convolutional layers, a ReLU activation, and a sigmoid gate:
$$
A_{fuse} = \sigma \left( \mathrm{Conv}_2 \left( \delta \left( \mathrm{Conv}_1 \left( [F_{cbam} \parallel G_{exp}] \right) \right) \right) \right).
$$
The final fused representation is generated in a residual gating form:
$$
F_{out} = F_{cbam} + \lambda \left( F_{cbam} \odot A_{fuse} \right),
$$
where $\odot$ denotes element-wise multiplication and $\lambda$ is a learnable scalar initialized to zero. 
This architectural design ensures that the RETFound-guided branch is initialized without glaucomatous specific bias, progressively learning to highlight clinically relevant retinal structures while suppressing spurious features. As a result, the impact of integrating this domain prior remains explicit, facilitating interpretability directly through the learned attention weights.

\subsection{Feature Fusion and Classification}

The three branches generate complementary feature embeddings, including the global representation $f_g$, the ROI structural representation $f_r$ and the dynamically localized representation $f_d$. These embeddings are concatenated to form the final feature vector $f_{final} = [f_g \parallel f_r \parallel f_d]$. The fused representation is then passed through a fully connected classifier to predict the glaucoma screening outcome:

\begin{equation}
y = Softmax(FC(f_{final}))
\end{equation}

The network outputs either three-class (\textit{Certain} vs. \textit{Suspect} vs. \textit{Normal}) or binary (\textit{referable} vs \textit{non-referable}) glaucoma predictions. By integrating global context, anatomical region features, and dynamically localized pathological cues under knowledge-guided attention, the proposed framework achieves robust and interpretable glaucoma screening across heterogeneous clinical datasets.

\section{Experiments}

\begin{figure}
    \centering
    \includegraphics[width=\columnwidth]{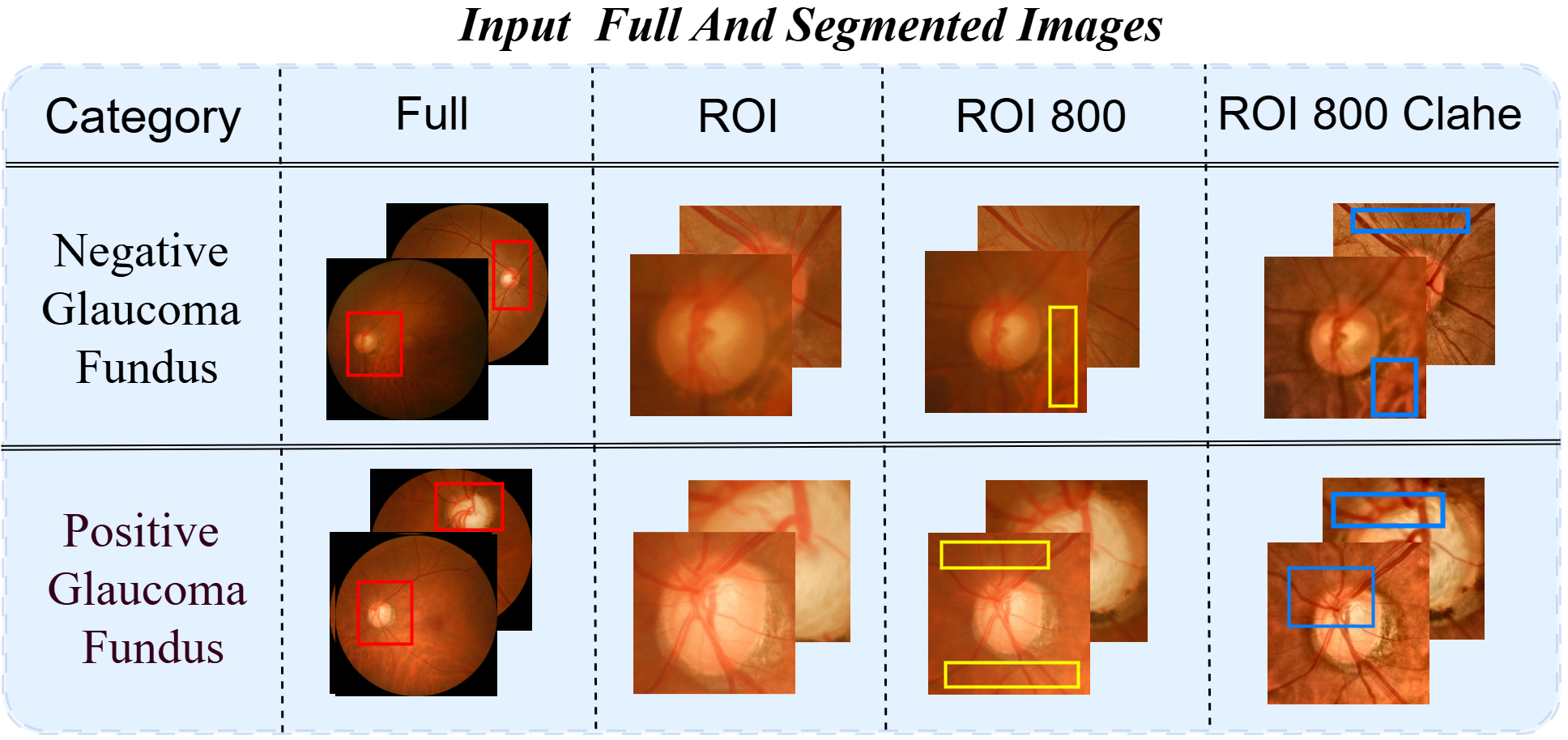}
    \caption{\textit{ROI 800} represents the enlarged region of interest around the optic disc. \textit{ROI 800 CLAHE} denotes the contrast-enhanced input image using Contrast Limited Adaptive Histogram Equalization (CLAHE). Red boxes indicate the ROI region, yellow boxes illustrate the enlarged view, and blue boxes highlight the contrast enhancement effect.}
    \label{Fig.1}
\end{figure}

\subsection{Dataset}
\label{Dataset}

Experiments are conducted on the Rotterdam EyePACS AIROGS dataset~\cite{r31}, a large-scale repository of color fundus photographs collected from multi-ethnic and geographically diverse populations. 
AIROGS selection is motivated by the scarcity of large-scale, publicly accessible cohorts with rigorous glaucoma annotations. As a universal benchmark in ophthalmic image analysis, AIROGS captures broad physiological and pathological variations, providing strong medical representativeness for real-world clinical scenarios.

Although originally labeled for binary referable glaucoma detection, we adopt a tri-class scheme: \textit{Negative}, \textit{Positive}, and \textit{Suspect}. The dataset is partitioned strictly at patient level, ensuring images from the same subject remain within a single split. The training set contains 36,803 images (27,519 Negative, 6,946 Positive, 2,338 Suspect), while validation and test sets contain 8,202 and 1,999 images, respectively, maintaining an approximate ratio of $0.75:0.19:0.06$, reflecting inherent class imbalance in clinical screening.

To address class imbalance, we employ standard dynamic data resampling and architectural integration of RETFound. By embedding robust retinal foundation knowledge as priors, the network is guided by learned anatomical representations
rather than relying solely on limited pathological samples. This domain-knowledge transfer compensates for numerical
scarcity of referable cases, enhancing discriminative capability and generalization on rare classes.

For cross-domain evaluation, we conduct experiments on the Standardized unseen Multi-Channel Dataset for Glaucoma SMDG-19~\cite{smdg}, the largest standardized glaucoma benchmark aggregating 19 public datasets from different
imaging devices and clinical centers, more details presented in Section.~\ref{comparison_baseline}. Seven representative datasets are selected to assess generalization under varying imaging conditions. This testing dataset is absolutely unseen, validating model generalizability.

Region of Interest (ROI) segmentation emphasizes the optic disc–cup complex. As illustrated in Figure~\ref{Fig.1}, Contrast Limited Adaptive Histogram Equalization (CLAHE) enhances \textit{ROI 800} images, serving as final model input.  CLAHE improves local contrast and enhances subtle anatomical structures, reducing illumination variation impact across imaging devices.

\subsection{Baseline Architectures}

\paragraph{\textbf{Baseline Models}}

To systematically evaluate the contribution of each architectural component, we compare the proposed method with several internal baseline variants representing the architectural evolution of the framework illustrated in Figure.~\ref{fig:concept}.

\textbf{Branch2NR}: Duel-branch architecture using ResNet152 backbone without attention.

\textbf{Patch5Model}: Multi-region baseline using ResNet152 extracting features from predefined patches without attention.

\textbf{Branch2CBAM}: Dual-branch architecture incorporating CBAM for channel–spatial feature interactions.

\textbf{Branch3CBAM}: Tri-branch architecture introducing DWM for adaptive pathological region localization.

\textbf{Branch3KECBAM}: Final proposed framework integrating KE-CBAM, injecting retinal anatomical priors from RETFound.

These variants allow progressive analysis of multi-scale representation learning, adaptive region localization, and knowledge-enhanced attention mechanisms.

\paragraph{\textbf{State-Of-The-Art Models}}

We compare with state-of-the-art glaucoma screening models to further evaluate its competitive performance.

\textbf{U-Nets-DenseNet}~\cite{li2022early}: Parallel U-Net branches segment vessels and determine ROI for DenseNet classification. 

\textbf{ResNet50-MaxVit}~\cite{zhao2023dual}:  Dual-branch network combining ResNet-50 and MaxViT for multi-scale local-global features with Swin Spatial Pyramid Pooling. 

\textbf{SA-GoogleNet}~\cite{alam2023segmentation}:  Two-step system using GoogleNet encoder for NV segmentation, then classification. 

\textbf{VisionDeep-AI}~\cite{joshi2024visiondeep}: Two-stage framework segmenting vessels via weighted bi-directional feature pyramid U-Net, then fusing features from original and masked images for four-class classification.

\textbf{AVS-DenseNet}~\cite{almeida2024enhancing}: Segmentation pipeline with DenseNet classifier, using novel filter to remove Optical Disc artifacts and Triangle Threshold algorithm. 

\textbf{Multi-GlaucNet}~\cite{xiong2025multi}: Multi-task model employing Pixel Shuffle and channel attention for parallel optic disc and vessel segmentation, feeding masks to detection network. 

\textbf{SegImgNet}~\cite{guan2026diffmcg}: Dual-branch model using U-Net segmentation, feeding entire images to Raw encoder and segmented images to ConvNeXt-based Segmented encoder with Segmentation-Guided Attention Block.

\subsection{Implementation Details} 
Experiments are conducted using the TensorFlow framework on the AIROGS dataset (refer to Section~\ref{Dataset}). Training is executed on a workstation equipped with four GeForce RTX 4090 GPUs, providing a total memory capacity of 192~GB (4$\times$48~GB). We employ the proposed 3-branch RCBAM framework utilizing a ResNet-152 backbone.

To address the challenges associated with class imbalance and the detection of high-difficulty samples in the three-class classification task, characterized by a class distribution of approximately $11:4:1$, we employ a dynamic data resampling strategy. Furthermore, comprehensive data augmentation techniques, including horizontal and vertical flipping, color jittering, and Gaussian blurring, are utilized with a probability of 0.5 ($P_{aug} = 0.5$) to improve model robustness.

\subsection{Training Parameters}
Model optimization is performed using the Adam optimizer ($\beta_1 = 0.9$, $\beta_2 = 0.999$). The mini-batch size is set to 16, distributed evenly across the four GPUs. The initial learning rate is configured to $5 \times 10^{-5}$. During training, when the validation performance saturates, the learning rate is halved iteratively until a minimum threshold of $2 \times 10^{-5}$ is reached. 

The network is trained for a maximum of 10,000 epochs. To ensure stable convergence and prevent overfitting, an early stopping mechanism is implemented with a patience of 3 epochs, triggered if the validation accuracy does not increase by a threshold of $\delta = 0.0001$. Training progress is monitored every 400 steps, with model checkpoints systematically archived every 5 iterations. Notably, no dropout layer is incorporated within the KE-CBAM branch; instead, model regularization and generalization are effectively controlled through the integration of the pretrained RETFound anatomical prior, residual gating mechanisms, zero-initialized fusion coefficients, and the aforementioned early stopping strategy.

\subsection{Evaluation Metrics}
\label{Evaluation_Metrics}

Classification performance is evaluated using six widely adopted metrics: Average Precision (AP), Area Under the ROC Curve (AUC), Accuracy (Acc), Sensitivity (Sen), Specificity (Spe), and F1-score. These metrics provide complementary perspectives on classification performance under different operating thresholds.

In addition, interpretability and feature representation are visualized in \textit{Top-6 Windows}, \textit{Grad-CAM++} saliency visualization and \textit{T-SNE} feature embedding, as described in Section~\ref{Visualization}.

\section{Results and Discussion}

\subsection{Quantitative Performance Comparison}

We conduct a comprehensive quantitative evaluation of the proposed framework from four complementary perspectives:
(a) tri-class classification to assess fine-grained pathological discrimination, 
(b) binary referral classification reflecting real-world clinical screening practice, 
(c) analyze the reliability of the ROI branch and compare the proposed model with the ablation without ROI-local branch and attention mechanism, 
(d) comparison with state-of-the-art methods together with cross-domain generalization analysis.

\begin{table*}[htbp]
\centering
\caption{Performance evaluation for tri-class and binary classification settings, with 5-fold means, standard deviations, and 95\% confidence intervals (0.95CI). The tri-class setting includes \textit{Negative}, \textit{Positive}, and \textit{Suspect}. The binary setting defines \textit{Non-referable} as \textit{Negative}, \textit{Referable} as (\textit{Positive} and \textit{Suspect})}
\label{tab:tri_binary}
\renewcommand{\arraystretch}{1.2} % 调整行高以适配期刊风格
% \scriptsize % 缩小字体以适应版心
\resizebox{\textwidth}{!}{
\setlength{\tabcolsep}{7.5mm}{

% -------- Top: Tri-class --------
\begin{tabular}{lcccccc}
\toprule
\multirow{2}{*}{\textbf{Model (3cls)}} & \multicolumn{6}{c}{\textbf{Evaluation Metrics}} \\ \cmidrule(lr){2-7}
 & AP & AUC & Acc & F1 & Sen & Spe \\ \hline
% Baseline 模型
Patch5Model (No AM) %
& \makecell{0.778 $\pm$ 0.014 \\ \scriptsize{[0.761, 0.796]}} 
& \makecell{0.958 $\pm$ 0.006 \\ \scriptsize{[0.951, 0.966]}} 
& \makecell{0.924 $\pm$ 0.008 \\ \scriptsize{[0.914, 0.934]}} 
& \makecell{0.726 $\pm$ 0.027 \\ \scriptsize{[0.692, 0.760]}} 
& \makecell{0.704 $\pm$ 0.034 \\ \scriptsize{[0.662, 0.747]}} 
& \makecell{0.927 $\pm$ 0.013 \\ \scriptsize{[0.911, 0.944]}} \\   \midrule

Branch3CBAM 
& \makecell{0.790 $\pm$ 0.014 \\ \scriptsize{[0.773, 0.807]}} 
& \makecell{0.963 $\pm$ 0.004 \\ \scriptsize{[0.958, 0.968]}} 
& \makecell{0.928 $\pm$ 0.004 \\ \scriptsize{[0.923, 0.932]}} 
& \makecell{0.770 $\pm$ 0.017 \\ \scriptsize{[0.749, 0.792]}}
& \makecell{0.767 $\pm$ 0.024 \\ \scriptsize{[0.738, 0.796]}} 
& \makecell{0.944 $\pm$ 0.006 \\ \scriptsize{[0.937, 0.951]}} 
 \\

 \midrule

\textbf{Branch3KECBAM}
& \makecell{0.806 $\pm$ 0.003 \\ \scriptsize{[0.803, 0.809]}} 
& \makecell{0.970 $\pm$ 0.001 \\ \scriptsize{[0.969, 0.971]}} 
& \makecell{0.934 $\pm$ 0.001 \\ \scriptsize{[0.933, 0.935]}} 
& \makecell{0.783 $\pm$ 0.006 \\ \scriptsize{[0.776, 0.790]}}
& \makecell{0.782 $\pm$ 0.008 \\ \scriptsize{[0.772, 0.792]}} 
& \makecell{0.949 $\pm$ 0.002 \\ \scriptsize{[0.947, 0.951]}} 
 \\ \midrule
\bottomrule
\end{tabular}}}

\vspace{3pt}

% -------- Bottom: Binary -------
\resizebox{\textwidth}{!}{
\setlength{\tabcolsep}{7.5mm}{
\begin{tabular}{lcccccc}
\hline
\multirow{2}{*}{\textbf{Models (binary)}} & \multicolumn{6}{c}{\textbf{Evaluation Metrics}} \\ \cline{2-7}
 & AP & AUC & Acc & F1 & Sen & Spe \\  \hline
Branch2NR (No ROI, AM) 
& \makecell{0.945 $\pm$ 0.013 \\ \scriptsize{[0.929, 0.960]}} 
& \makecell{0.978 $\pm$ 0.005 \\ \scriptsize{[0.972, 0.985]}} 
& \makecell{0.925 $\pm$ 0.014 \\ \scriptsize{[0.908, 0.943]}} 
& \makecell{0.854 $\pm$ 0.016 \\ \scriptsize{[0.834, 0.873]}} 
& \makecell{0.884 $\pm$ 0.069 \\ \scriptsize{[0.798, 0.970]}} 
& \makecell{0.939 $\pm$ 0.040 \\ \scriptsize{[0.889, 0.988]}} \\   \midrule

Patch5Model (No AM) 
& \makecell{0.952 $\pm$ 0.008 \\ \scriptsize{[0.942, 0.962]}} 
& \makecell{0.980 $\pm$ 0.004 \\ \scriptsize{[0.975, 0.985]}} 
& \makecell{0.931 $\pm$ 0.013 \\ \scriptsize{[0.915, 0.948]}} 
& \makecell{0.842 $\pm$ 0.039 \\ \scriptsize{[0.794, 0.891]}} 
& \makecell{0.761 $\pm$ 0.071 \\ \scriptsize{[0.673, 0.848]}} 
& \makecell{0.986 $\pm$ 0.006 \\ \scriptsize{[0.978, 0.994]}} \\   \midrule

Branch2CBAM (No DWM) 
& \makecell{0.948 $\pm$ 0.009 \\ \scriptsize{[0.937, 0.959]}} 
& \makecell{0.979 $\pm$ 0.003 \\ \scriptsize{[0.975, 0.982]}} 
& \makecell{0.936 $\pm$ 0.003 \\ \scriptsize{[0.931, 0.940]}} 
& \makecell{0.869 $\pm$ 0.010 \\ \scriptsize{[0.856, 0.882]}} 
& \makecell{0.875 $\pm$ 0.050 \\ \scriptsize{[0.813, 0.937]}} 
& \makecell{0.955 $\pm$ 0.016 \\ \scriptsize{[0.936, 0.975]}} \\   \midrule

Branch3CBAM 
& \makecell{0.959 $\pm$ 0.001 \\ \scriptsize{[0.958, 0.960]}} % AP
& \makecell{0.984 $\pm$ 0.001 \\ \scriptsize{[0.983, 0.984]}} % AUC
& \makecell{0.946 $\pm$ 0.001 \\ \scriptsize{[0.945, 0.947]}} % Acc
& \makecell{0.890 $\pm$ 0.002 \\ \scriptsize{[0.888, 0.892]}} % F1
& \makecell{0.896 $\pm$ 0.006 \\ \scriptsize{[0.889, 0.904]}} % Sen
& \makecell{0.962 $\pm$ 0.003 \\ \scriptsize{[0.959, 0.965]}} \\ \midrule

\textbf{Branch3KECBAM} 
& \makecell{0.958 $\pm$ 0.003 \\ \scriptsize{[0.955, 0.962]}}
& \makecell{0.982 $\pm$ 0.001 \\ \scriptsize{[0.981, 0.983]}}
& \makecell{0.950 $\pm$ 0.002 \\ \scriptsize{[0.948, 0.953]}}
& \makecell{0.897 $\pm$ 0.004 \\ \scriptsize{[0.893, 0.902]}}
& \makecell{0.885 $\pm$ 0.010 \\ \scriptsize{[0.8723, 0.898]}}
& \makecell{0.971 $\pm$ 0.005 \\ \scriptsize{[0.965, 0.978]}} \\ \midrule
\end{tabular}}}

\end{table*}

\subsubsection{Tri-class Classification Performance}

To evaluate the capability of the proposed model in capturing subtle pathological variations, we first report results under a tri-class setting comprising \textit{Negative}, \textit{Suspect}, and \textit{Positive} glaucoma. As shown in Table~\ref{tab:tri_binary}, the proposed \textit{Branch3KECBAM} is compared with the baseline \textit{Patch5Model} and the \textit{Branch3CBAM}.

The tri-class task remains challenging as \textit{Suspect} cases exhibit transitional structural changes between
normal and glaucomatous patterns. Despite this, \textit{Branch3KECBAM} achieves the best performance across all
metrics. Compared with \textit{Branch3CBAM}, it improves AP from 0.790 to 0.806, AUC from 0.963 to 0.970, and accuracy
from 0.928 to 0.934. Critically, F1-score and sensitivity increase from 0.770 to 0.783 and from 0.767 to 0.782,
respectively, indicating stronger identification of clinically relevant cases without sacrificing specificity from 0.944 to 0.949.

Compared with \textit{Patch5Model}, \textit{Branch3KECBAM} shows gains of 0.028 in AP, 0.012 in AUC, 0.010 in accuracy, 0.057 in F1-score, and 0.078 in sensitivity with reduced cross-validation variance. These improvements demonstrate that KE-CBAM provides more discriminative representations by incorporating retinal anatomical priors into
the attention mechanism, enhancing discrimination of ambiguous \textit{Suspect} cases.

\subsubsection{Binary Clinical Referral Decision}

Although the tri-class setting enables a fine-grained assessment of disease severity, practical glaucoma screening is primarily concerned with binary referral decisions. Therefore, the \textit{Suspect} and \textit{Positive} categories are merged into a single \textit{Referable} class, whereas \textit{Negative} samples are treated as \textit{Non-referable} cases.

Table~\ref{tab:tri_binary} shows all models achieve strong performance with differing error profiles. \textit{Patch5Model} obtains high AP 0.952, AUC 0.980, and specificity 0.986, but limited sensitivity 0.761, suggesting conservative predictions that miss referable cases. Introducing CBAM improves balance: \textit{Branch2CBAM} increases sensitivity to 0.875 and F1 to 0.869, while \textit{Branch3CBAM} achieves sensitivity of 0.896 and F1 of 0.890, indicating adaptive attention captures pathological cues.

The third branch in \textit{Branch3CBAM} yields AP of 0.959, AUC of 0.984, and sensitivity of 0.896, providing complementary regional information. The proposed \textit{Branch3KECBAM} further improves decision balance, achieving highest accuracy 0.950, F1-score 0.897, and specificity 0.971. Although AP and AUC are marginally lower than \textit{Branch3CBAM}, the gains suggest a more favorable trade-off between identifying referable cases and reducing false referrals.

These results demonstrate that knowledge-enhanced attention contributes to clinically consistent screening decisions by emphasizing diagnostically relevant structures while suppressing irrelevant responses.

\begin{figure}
\centering
\begin{minipage}{0.49\linewidth}
    \centering
    \includegraphics[width=\linewidth]{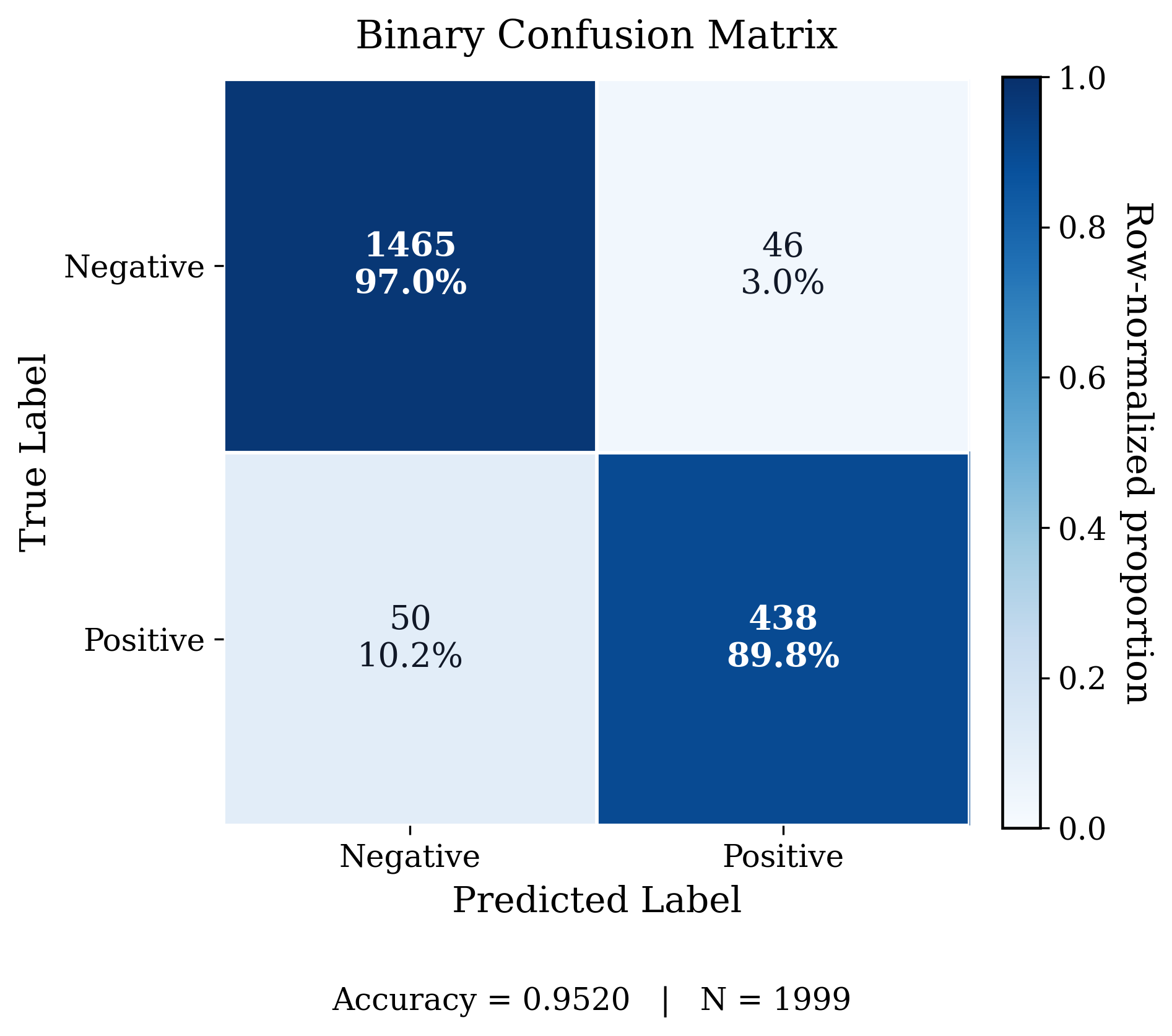}
\end{minipage}\hfill
\begin{minipage}{0.49\linewidth}
    \centering
    \includegraphics[width=\linewidth]{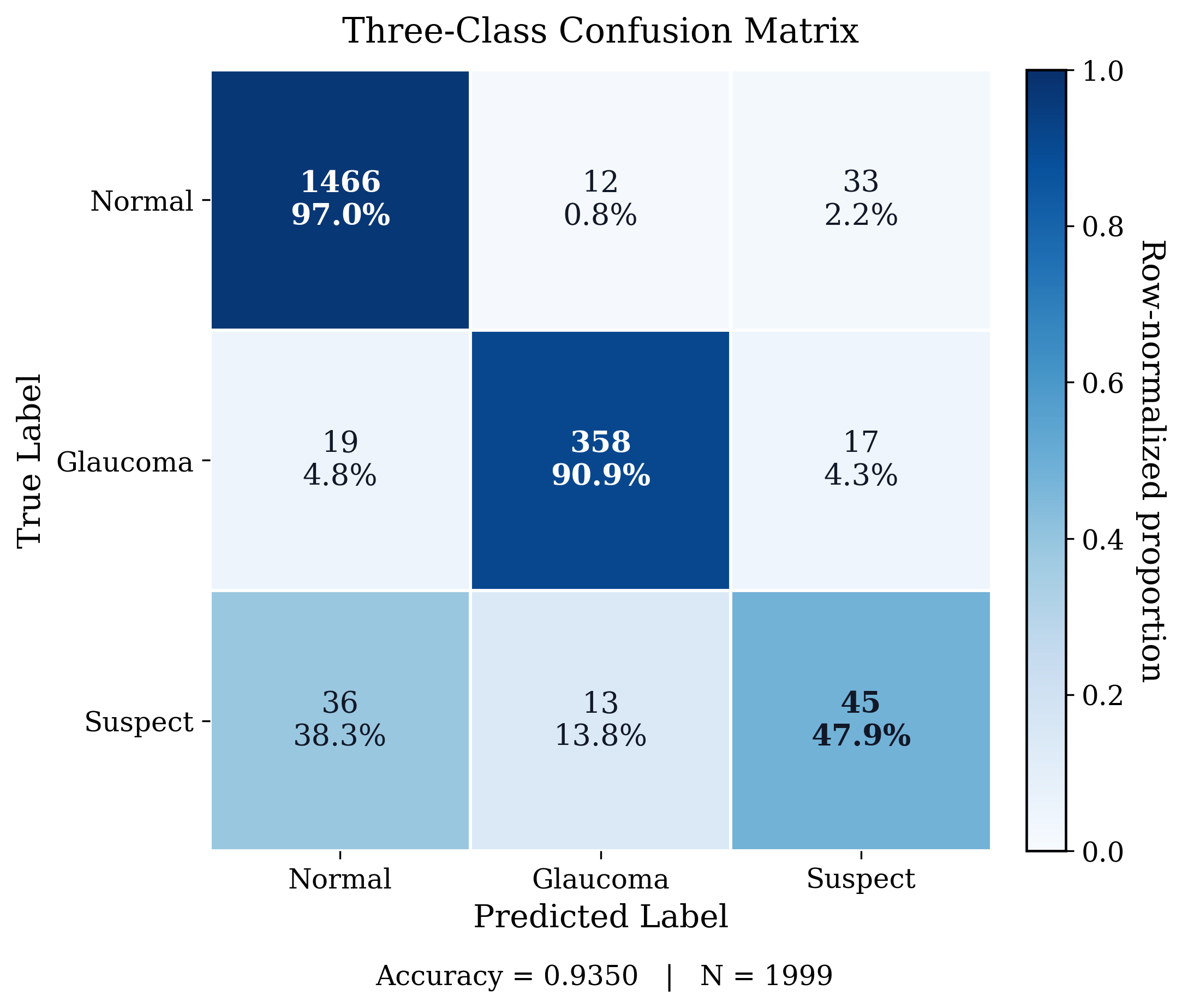}
\end{minipage}

\caption{Row-normalized confusion matrices of the proposed framework under (a) tri-class (Negative/Suspect/Positive) and (b) binary (non-referable vs referable) settings.}
\label{fig4}
\end{figure}

Figure~\ref{fig4} shows confusion matrices for both the binary and tri-class tasks from the 5-fold cross-validation, indicating most tri-class errors arise from ambiguous \textit{Suspect} cases, reflecting gradual glaucoma progression. Merging into \textit{Referable} aligns the decision boundary with practical screening, yielding more stable referral outcomes.

\subsubsection{Reliability of ROI and Ablation}
\label{sec:roi_ablation}

\begin{figure}
    \centering
    \begin{minipage}{\linewidth}
        \centering

        % -------- Row 1: Full Fundus Images --------
        \begin{subfigure}[t]{0.19\linewidth}
            \centering
            \includegraphics[width=\linewidth]{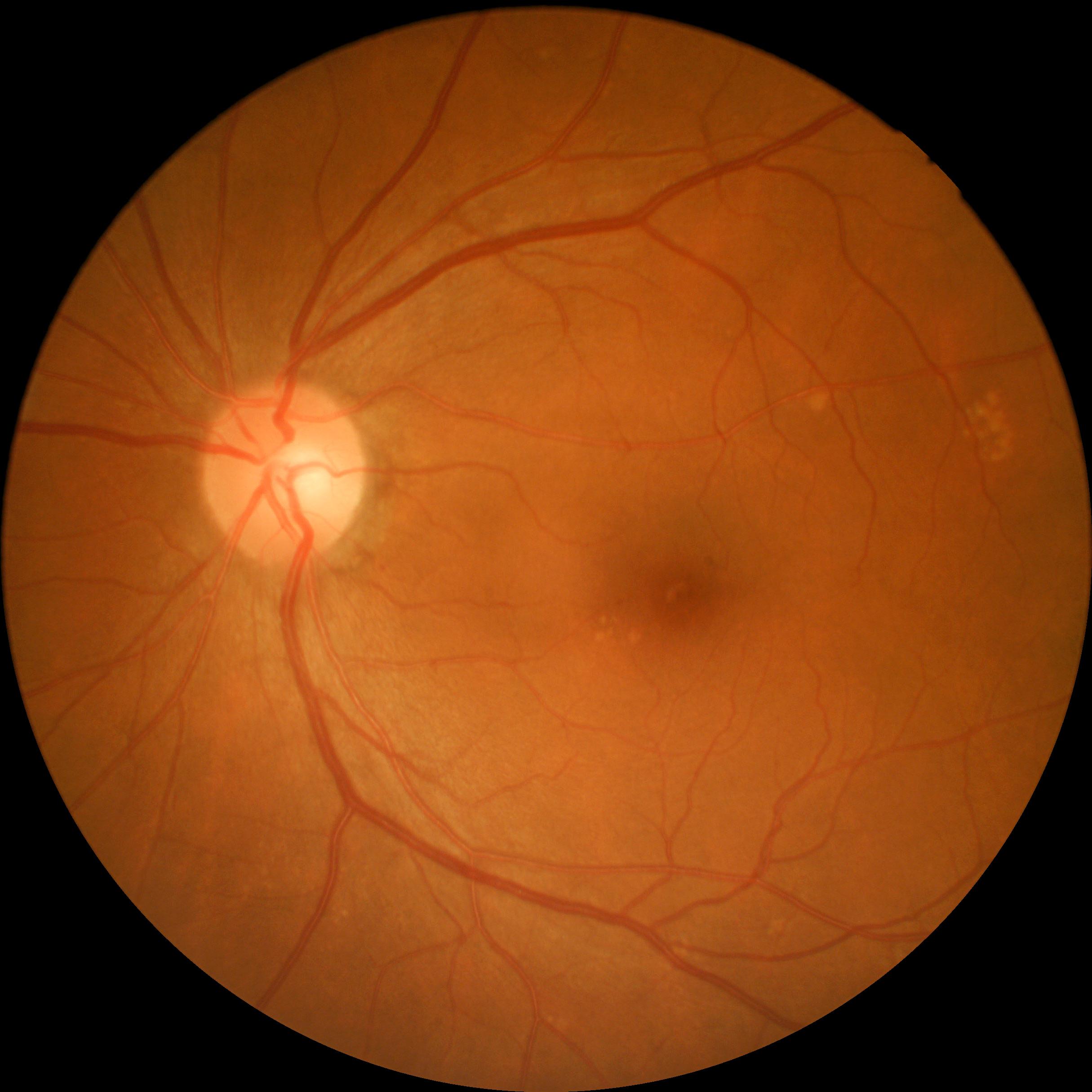}
            \caption{}
        \end{subfigure}\hfill
        \begin{subfigure}[t]{0.19\linewidth}
            \centering
            \includegraphics[width=\linewidth]{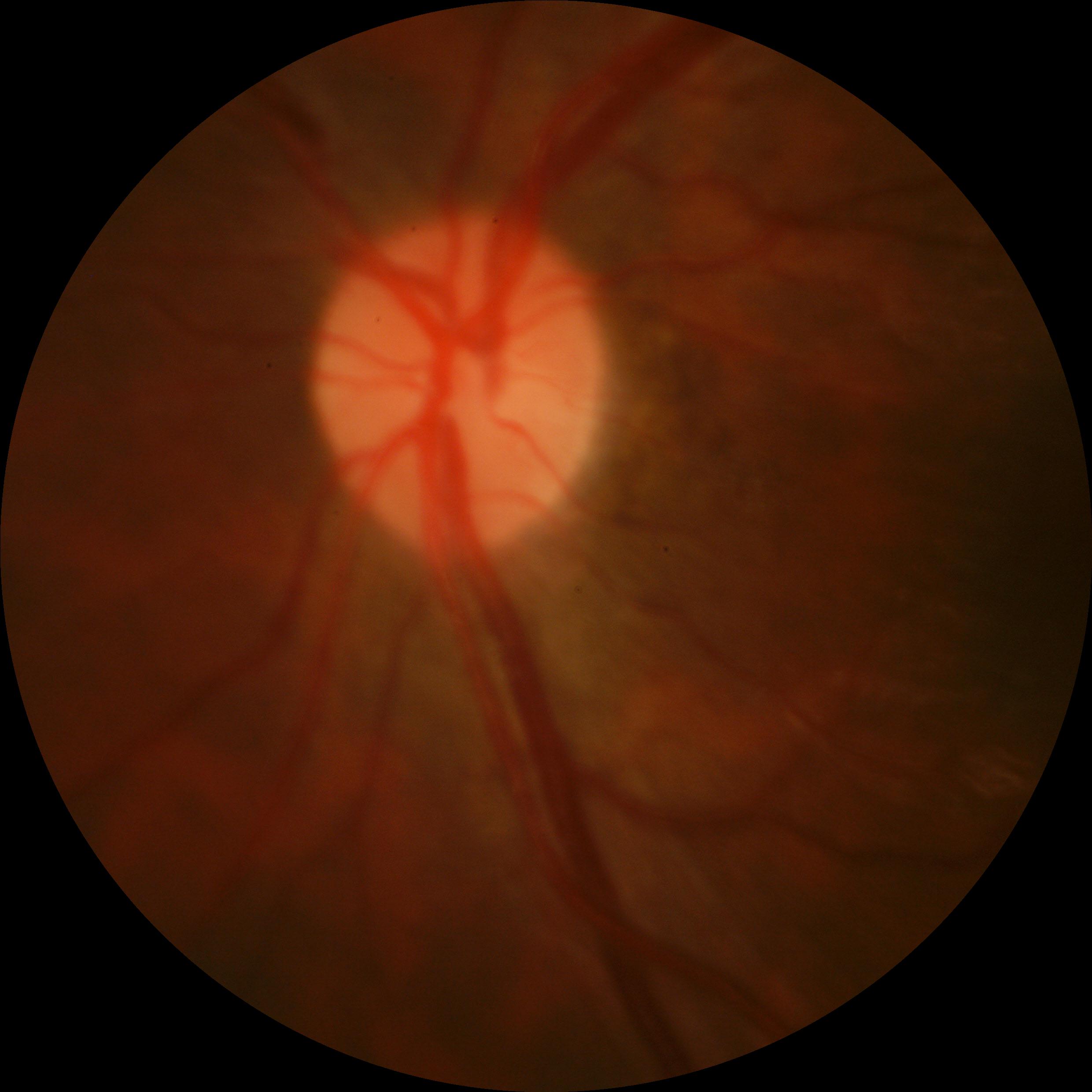}
            \caption{}
        \end{subfigure}\hfill
        \begin{subfigure}[t]{0.19\linewidth}
            \centering
            \includegraphics[width=\linewidth]{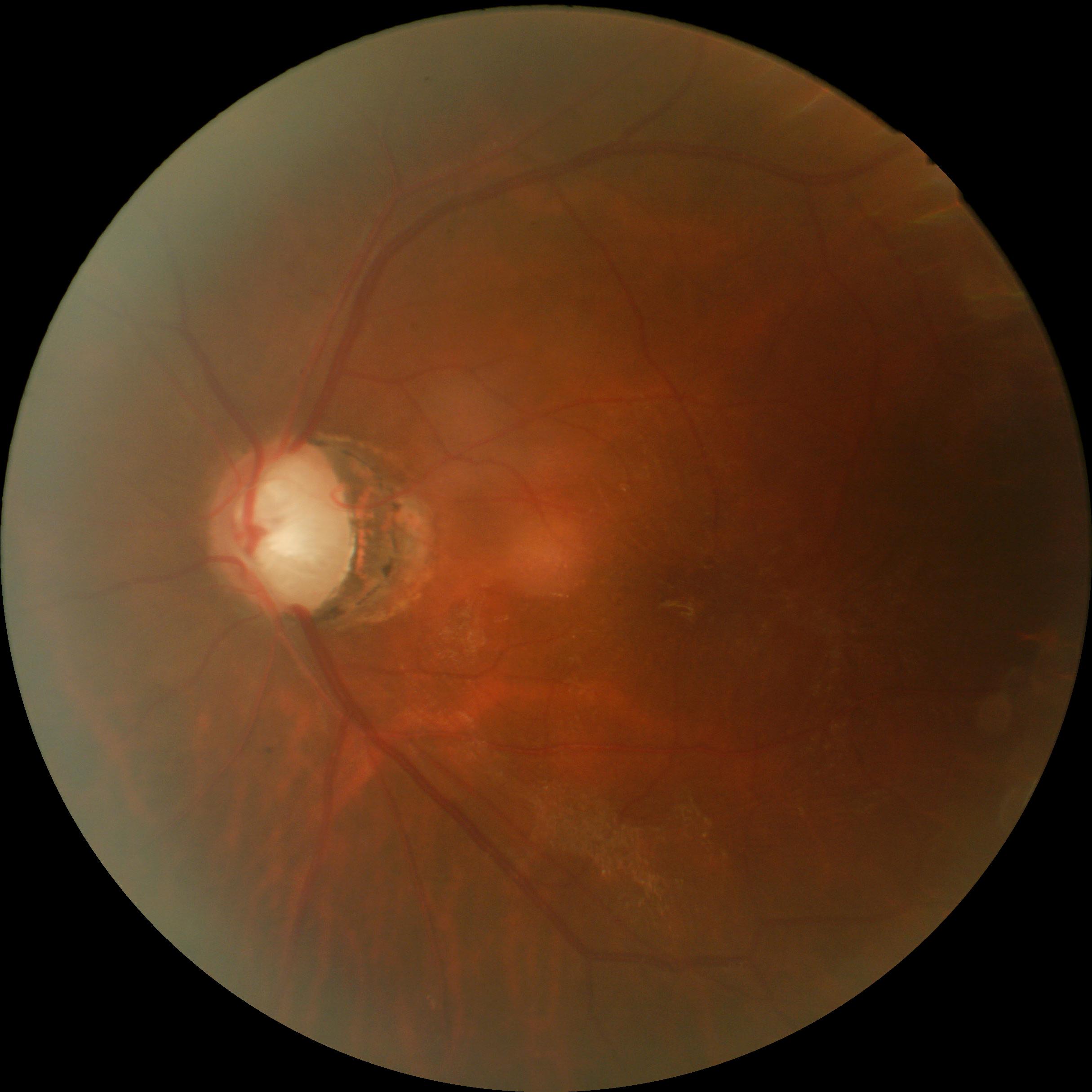}
            \caption{}
        \end{subfigure}\hfill
        \begin{subfigure}[t]{0.19\linewidth}
            \centering
            \includegraphics[width=\linewidth]{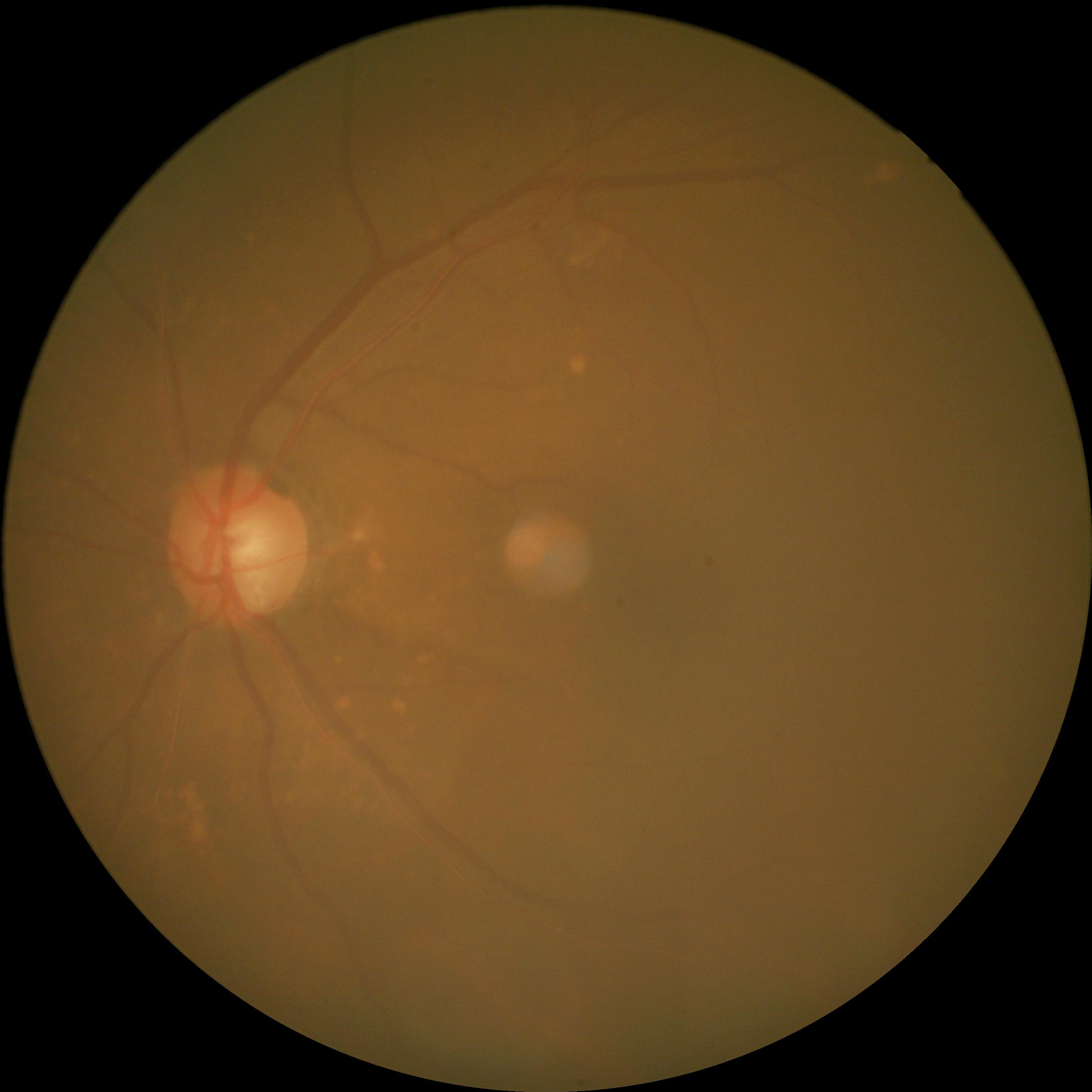}
            \caption{}
        \end{subfigure}\hfill
        \begin{subfigure}[t]{0.19\linewidth}
            \centering
            \includegraphics[width=\linewidth]{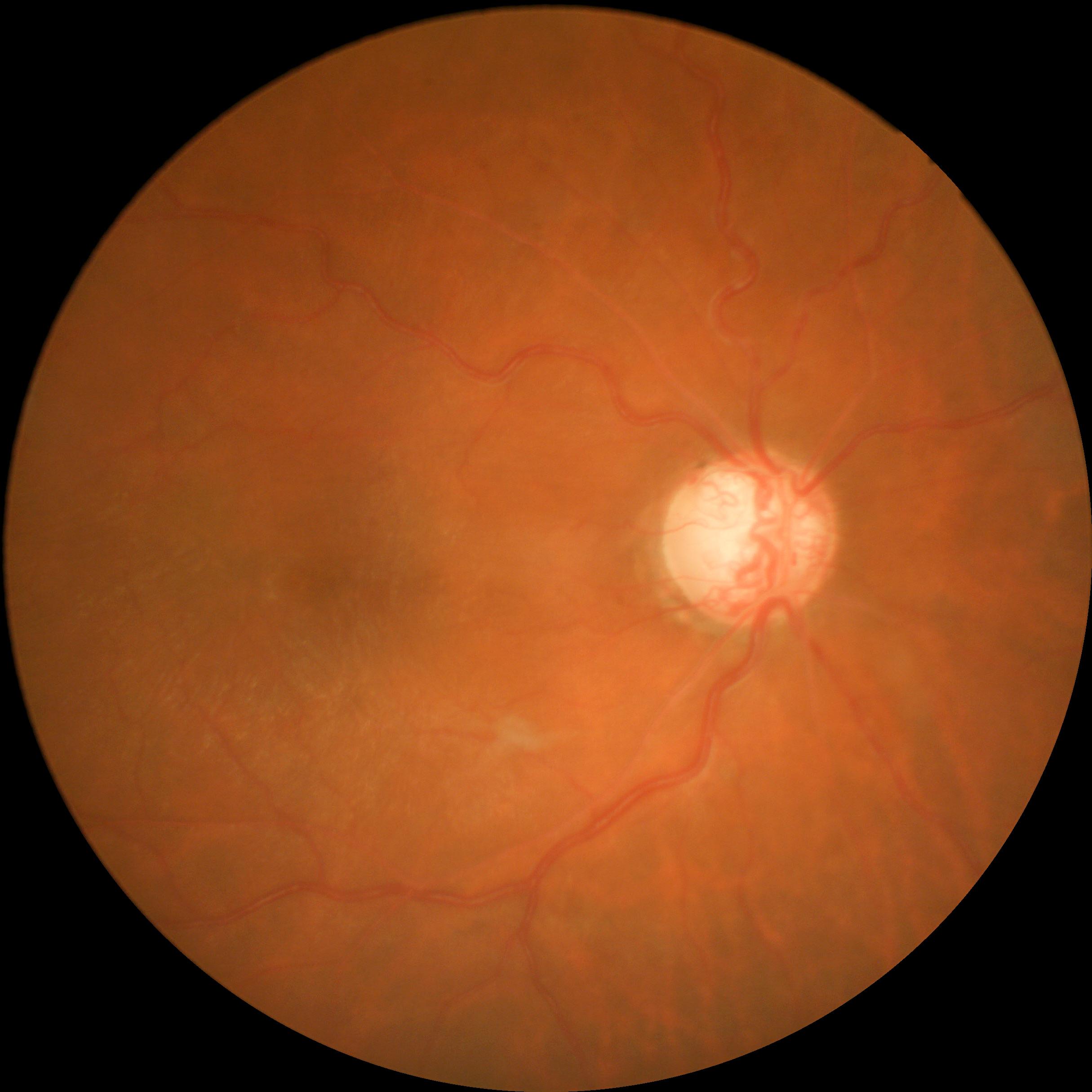}
            \caption{}
        \end{subfigure}

        \vspace{2mm}

        % -------- Row 2: ROI Crops --------
        \begin{subfigure}[t]{0.19\linewidth}
            \centering
            \includegraphics[width=\linewidth]{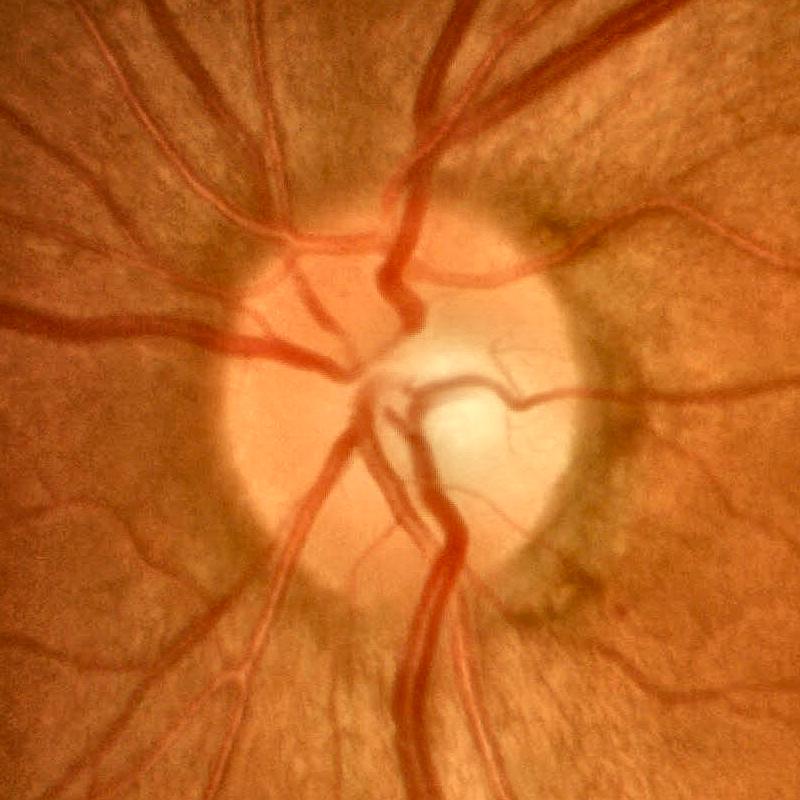}
            \caption{}
        \end{subfigure}\hfill
        \begin{subfigure}[t]{0.19\linewidth}
            \centering
            \includegraphics[width=\linewidth]{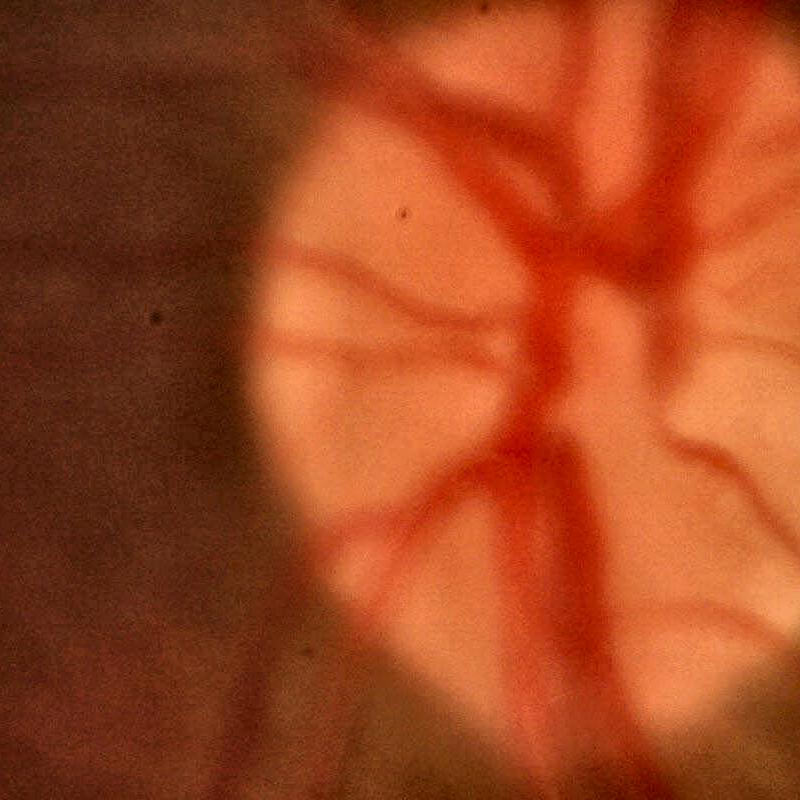}
            \caption{}
        \end{subfigure}\hfill
        \begin{subfigure}[t]{0.19\linewidth}
            \centering
            \includegraphics[width=\linewidth]{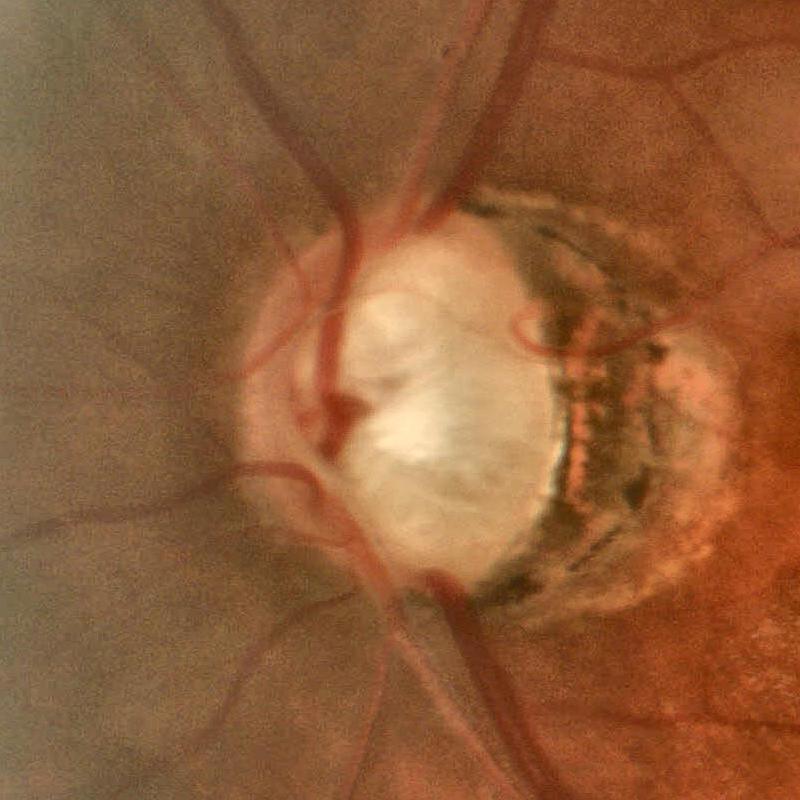}
            \caption{}
        \end{subfigure}\hfill
        \begin{subfigure}[t]{0.19\linewidth}
            \centering
            \includegraphics[width=\linewidth]{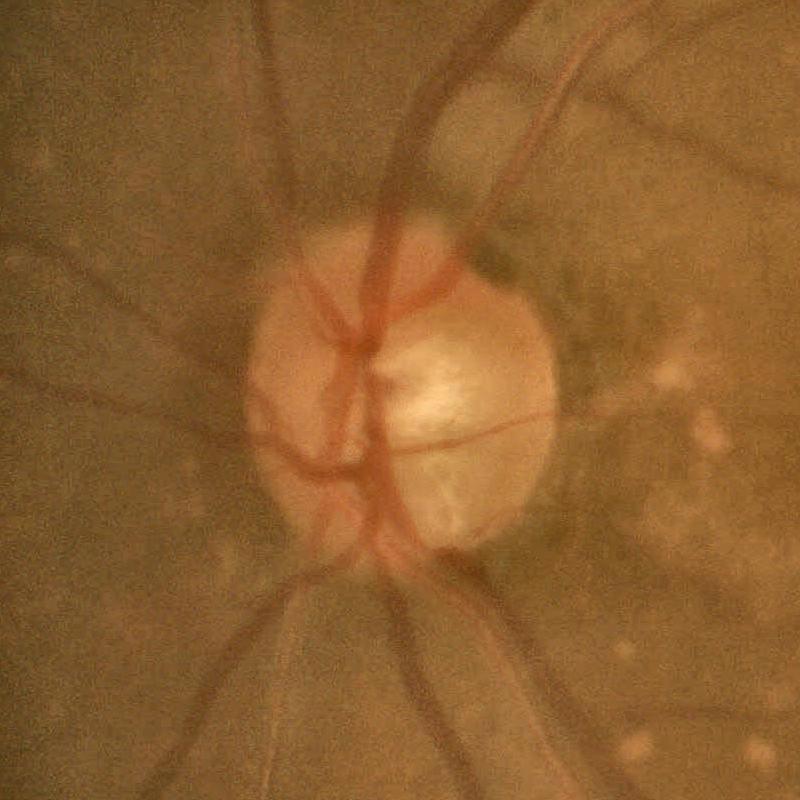}
            \caption{}
        \end{subfigure}\hfill
        \begin{subfigure}[t]{0.19\linewidth}
            \centering
            \includegraphics[width=\linewidth]{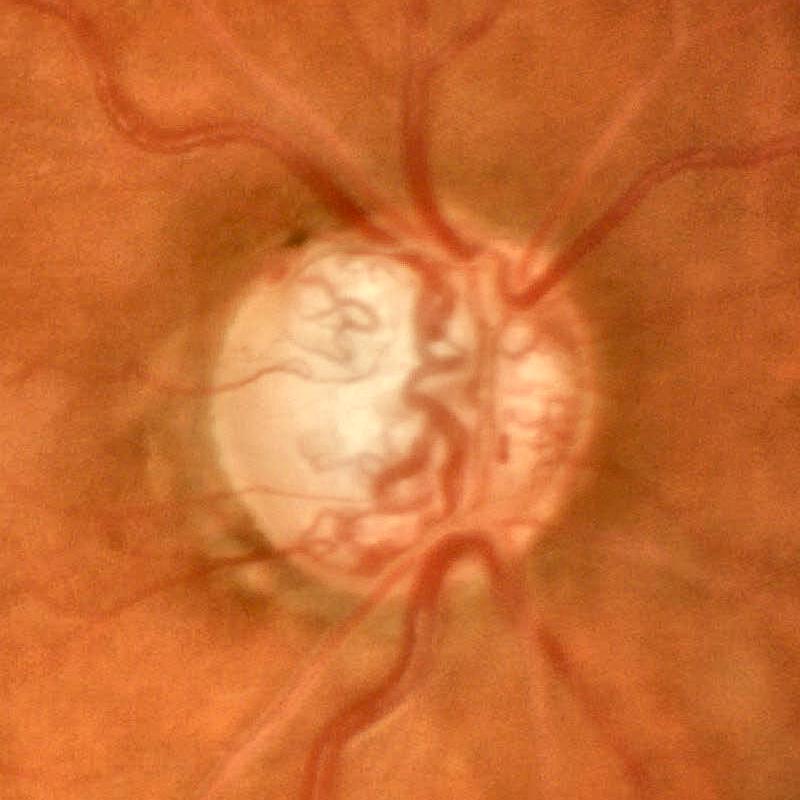}
            \caption{}
        \end{subfigure}

    \end{minipage}
    \caption{Representative examples of optic disc ROI extraction under diverse real-world imaging conditions. Top row: (a) standard high-quality image, (b) severe out-of-focus with blurred boundaries, (c) low-resolution with local artifacts, (d) severely underexposed with dim illumination, (e) low-contrast with weak vessel patterns. Bottom row (f-j): corresponding ROI crops demonstrating consistent optic disc localization across all conditions.}
    \label{fig:roi_reliability}
\end{figure}

To address concerns about optic disc segmentation reliability under varying image quality, we conducted visual analysis on representative fundus images covering five clinical conditions in Figure~\ref{fig:roi_reliability}. Despite substantial quality variations including blur, low resolution, artifacts, underexposure, and low contrast, the ROI crops consistently localize and center the optic disc, while CLAHE enhancement further improves local contrast for feature extraction.

Despite substantial quality variations, the ROI crops consistently localize and center the optic disc across all cases. The segmentation pipeline remains stable under blur, low resolution, artifacts, underexposure, and low contrast, while CLAHE enhancement further improves local contrast for feature extraction. Table.~\ref{tab:tri_binary} quantitatively validate the ROI-Local branch contribution. Introducing the ROI branch to \textit{Branch2NR}, \textit{Branch3CBAM} increases AP from 0.945 to 0.959, AUC from 0.978 to 0.984, accuracy from 0.925 to 0.946, and F1-score from 0.854 to 0.890. Sensitivity improves from 0.884 to 0.896, while specificity increases from 0.939 to 0.962. These gains indicate the ROI branch provides discriminative anatomical information from the optic disc-cup region.

Furthermore, reduced performance variability in 5-fold cross-validation confirms that segmentation preprocessing does not introduce unstable noise. The ROI-Local branch provides stable, clinically relevant structural cues that enhance global image representation, supporting its reliability in the proposed \textit{Branch3KECBAM}.

\subsubsection{Comparison with State-of-the-Art Methods}
\label{comparison_baseline}

To comprehensively evaluate the competitiveness and generalizability of the proposed framework, we compare our final model, \textbf{ResNet152-KE-CBAM}, with several recent state-of-the-art glaucoma detection methods. The comparative results on the large-scale AIROGS dataset and the multi-domain SMDG-19 benchmark are summarized in Table~\ref{tab:combined_evaluation}.

\paragraph{\textbf{Diagnostic Performance on the AIROGS Dataset}} 
As presented in Table~\ref{tab:combined_evaluation}, the proposed model achieves an accuracy of 94.55\% and an AUC of 98.52\% on the AIROGS dataset, consistently outperforming recent alternative approaches. Notably, the model yields the highest specificity (96.70\%). This metric is particularly critical in large-scale clinical screening programs, as a reduced false-positive rate significantly alleviates the unnecessary burden on healthcare systems. These substantial improvements can be directly attributed to the proposed knowledge-enhanced attention mechanism. By incorporating retinal anatomical priors derived from the RETFound foundation model, KE-CBAM explicitly guides the network's attention toward pathologically relevant structures, preventing the model from relying solely on data-driven feature learning.

\begin{table}[!t]
\centering
\caption{Comprehensive performance evaluation: \textbf{Part I} compares our proposed method with recent state-of-the-art methods on the EyePACS dataset. \textbf{Part II} evaluates the generalization capability across seven public datasets from the SMDG-19~\cite{smdg} in baseline comparisons~\cite{shah2025optiguard}.}
\label{tab:combined_evaluation}
\renewcommand{\arraystretch}{1.2}

% ---------------- Part I: EyePACS Dataset ----------------
\setlength{\tabcolsep}{4pt}
\resizebox{\linewidth}{!}{%
\begin{tabular}{lccccc}
\toprule
\multicolumn{6}{c}{\textbf{Part I: Performance comparison on EyePACS Dataset}} \\
\midrule
\textbf{Author} & \textbf{Method} & \textbf{Acc (\%)} & \textbf{Sen (\%)} & \textbf{Spe (\%)} & \textbf{AUC (\%)} \\
\midrule
Li and Liu (2022)~\cite{li2022early} & U-Nets-DenseNet & 83.40 & 82.30 & 84.60 & 91.50\\
Zhao et al. (2023)~\cite{zhao2023dual} & ResNet50-MaxViT & 91.60 & 92.50 & 90.70 & 97.00\\
Alam et al. (2023)~\cite{alam2023segmentation} & SA-GoogleNet & 90.00 & 90.00 & 90.10 & 96.10 \\
Joshi et al. (2024)~\cite{joshi2024visiondeep} & VisionDeep-AI & 91.40 & 93.00 & 89.90 & 97.00 \\
Almeida et al. (2024)~\cite{almeida2024enhancing} & AVS-DenseNet & 77.80 & 85.90 & 69.80 & 85.00 \\
Xiong et al. (2025)~\cite{xiong2025multi} & Multi-GlaucNet & 89.70 & \textbf{94.90} & 84.40 & 96.50 \\
Guan et al. (2026)~\cite{guan2026diffmcg} & SegImgNet & 94.00 & \textbf{94.90} & 93.40 & 98.50 \\
\midrule
\textbf{Proposed} & \textbf{ResNet152-KE-CBAM} & \textbf{95.04} & 88.52 & \textbf{97.14} & \textbf{98.22} \\
\bottomrule
\end{tabular}%
}

\vspace{0.8em}

% ---------------- Part II: SMDG-19 Datasets ----------------
\setlength{\tabcolsep}{3pt}
\resizebox{\linewidth}{!}{%
\begin{tabular}{lccccccc}
\toprule
\multicolumn{8}{c}{\textbf{Part II: Accuracy comparison on SMDG-19 Datasets}} \\
\midrule
\textbf{Model / Dataset} & \textbf{REFUGE} & \textbf{G1020} & \textbf{PAPILA} & \textbf{ORIGA} & \textbf{BEH} & \textbf{OIA-ODIR} & \textbf{sjchoi-HRF} \\
\midrule
ResNet50            & 0.930 & 0.587 & 0.850 & \textbf{0.801} & 0.718 & 0.903 & 0.829 \\
EfficientNet-b2     & 0.938 & 0.548 & \textbf{0.883} & 0.772 & 0.781 & 0.863 & \textbf{0.851} \\
Inception-Resnet-v2 & 0.826 & 0.522 & 0.750 & 0.722 & 0.666 & 0.806 & 0.787 \\
\midrule
\textbf{ResNet152-KE-CBAM} & \textbf{0.959} & \textbf{0.697} & 0.763 & 0.794 & \textbf{0.813} & \textbf{0.921} & 0.803 \\
\bottomrule
\end{tabular}%
}
\end{table}

\paragraph{\textbf{Cross-Domain Generalization on the SMDG-19 Benchmark}} 

Performance variations across the SMDG-19 subsets offer a practical measure of the model's sensitivity to domain shifts. These fluctuations stem directly from the extreme clinical and physical heterogeneities inherent in the assembled datasets, demonstrating the framework's comprehensive generalizability in real-world environments.

The benchmark inherently captures deep demographic and regional disparities, which dictate natural variations in fundus pigmentation. The collection spans distinct regional cohorts, encompassing European (G1020~\cite{bajwa2020g1020}), East Asian (ORIGA~\cite{zhang2010origa}, sjchoi86-HRF~\cite{arias2024artificial}), and South Asian (BEH~\cite{zhihu_beh_dataset_2024}) populations. Despite the visible differences in retinal appearance across these ethnic groups, the proposed framework sustains stable diagnostic accuracy, highlighting its independence from population biases. 

This robustness extends beyond demographics into the technical challenges of image acquisition. Hardware heterogeneity is particularly evident in the OIA-ODIR~\cite{nkicsl_oia_odir_2019} subset, which aggregates data from multiple clinical centers using various commercial cameras, including \textit{Canon, Zeiss, Kowa}, which causes drastic inconsistencies in resolution and color spaces. PAPILA~\cite{kovalyk2022papila} relies on non-mydriatic cameras \textit{Topcon TRC-NW400} that frequently produce underexposed images. In contrast, G1020 captures mydriatic images without imposing spatial constraints, resulting in highly decentralized optic discs. 

Alongside hardware inconsistencies, the age distribution and accompanying ocular conditions introduce significant clinical noise. The PAPILA dataset features older patients between 47 and 79 years, while subsets like OIA-ODIR and sjchoi86-HRF include non-glaucomatous pathologies, such as cataracts, which obscure underlying retinal structures. Despite, by integrating the DWM with KE-CBAM, the  system consistently isolates invariant glaucomatous features across varying ethnicities, camera models, and patient age groups, proving its reliable generalization capacity for complex clinical applications.

\subsection{Model Complexity and Computational Cost}
\label{subsec:complexity}

This section aims to quantify the computational cost of the proposed three architectural design and to clarify the performance-efficiency trade-off. Three variants are evaluated, \textbf{3branch}, \textbf{3branch-cbam}, and \textbf{3branch-rcbam}, and their complexity is compared in terms of trainable parameters, FLOPs, practical inference efficiency, latency and FPS, as summarized in Table~\ref{tab:complexity}.

\begin{table}[t]
\centering
\setlength{\tabcolsep}{10pt}
\renewcommand{\arraystretch}{1.15}
\caption{Model complexity and computational efficiency comparison across three architectural variants. Values are reported as mean $\pm$ standard deviation, with 95\% confidence intervals shown in brackets for time and FPS. FLOPs are measured with input size $299 \times 299 \times 3$.}
\label{tab:complexity}
\resizebox{\columnwidth}{!}{%
\begin{tabular}{lccccc}
\toprule
\textbf{Model} & \textbf{Params (M)} & \textbf{FLOPs (G)} & \textbf{Time (ms)} & \textbf{FPS} & \textbf{F1 (\%)} \\
\midrule
3branch (Baseline)  &
60.75 &
171.73 &
$20.93 \pm 3.50$  &
$49.13 \pm 12.70$  &
$84.2 \pm 3.9 $ \\
3branch-cbam &
67.34 &
172.06 &
$38.51 \pm 0.18$  &
$25.97 \pm 0.15$  &
$89.03 \pm 0.20$\\
\textbf{3branch-rcbam} &
83.59 &
200.66 &
$62.56 \pm 0.23$  &
$15.99 \pm 0.07$  &
$89.70 \pm 0.36$ \\
\midrule
\textit{Proposed vs. baseline} &
+37.7\% &
+16.8\% &
+198.8\% &
-67.4\% &
+5.6\% \\
\bottomrule
\end{tabular}%
}
\end{table}

\subsubsection{Computational Complexity Analysis}
\label{subsubsec:complexity_static}

Table~\ref{tab:complexity} reports the static complexity of the three variants. The baseline \textbf{3branch}
establishes a reference with 60.75M parameters and 171.73G FLOPs. \textbf{3branch-cbam} increases parameters to 67.34M
with negligible FLOPs change (172.06G), indicating CBAM's cost is dominated by lightweight operations.

The proposed \textbf{3branch-rcbam} exhibits 83.59M parameters and 200.66G FLOPs, corresponding to +37.7\% parameters
and +16.8\% FLOPs relative to baseline. This structured increase reflects the computational budget required for
RETFound-guided refined attention to integrate global priors with local multi-scale features.

\subsubsection{Practical Inference Efficiency}
\label{subsubsec:complexity_runtime}

Beyond static metrics, Table~\ref{tab:complexity} provides operational efficiency via latency and throughput. The
baseline \textbf{3branch} achieves $20.93 \pm 3.50$ ms with $49.13 \pm 12.70$ FPS. Despite negligible FLOPs increase,
\textbf{3branch-cbam} nearly doubles runtime to $38.51 \pm 0.18$ ms and $25.97 \pm 0.15$ FPS, revealing memory-bound
  overhead of attention modules.

\textbf{3branch-rcbam} further increases latency to $62.56 \pm 0.23$ ms ($15.99 \pm 0.07$ FPS), representing +198.8\%
latency and -67.4\% throughput relative to baseline. The tight variability indicates stable, predictable runtime behavior desirable for deployment.

\subsubsection{Performance Gains and Trade-off}
\label{subsubsec:complexity_tradeoff}

The proposed \textbf{3branch-rcbam} improves F1-score from $84.2 \pm 3.9\%$ to $88.9 \pm 0.2\%$, yielding +5.6\% gain with markedly reduced variability. This stability indicates that RETFound-guided refined attention consistently enhances robustness across cross-validation splits by injecting global anatomical priors, reducing reliance on fluctuating fine-grained cues.

The core trade-off is clear: \textbf{3branch-rcbam} attains substantial, consistent F1 improvement at the cost of increased computational overhead. In clinical screening workflows prioritizing diagnostic accuracy over throughput, this trade-off is practically justified. Our explicit reporting of both efficiency and performance enables transparent model selection based on hardware constraints and application requirements.

\begin{table}[!t]
\centering
\caption{Performance comparison between the Patch5Model and proposed Branch3KECBAM under three degradation levels (lvl = 0, 1, 2) for blur, JPEG compression, and Gaussian noise. Values are reported as mean $\pm$ standard deviation across 5 folds.}
\label{tab:robustness}
\setlength{\tabcolsep}{6pt}
\renewcommand{\arraystretch}{1.15}
\resizebox{\linewidth}{!}{%
\begin{tabular}{llccccc}
\toprule
\textbf{Degradation} & \textbf{Model} & \textbf{Level} & \textbf{Acc.} & \textbf{Sens.} & \textbf{F1} & \textbf{AUC} \\
\midrule
\multirow{6}{*}{Blur}
 & baseline & 0 & $0.9373 \pm 0.0057$ & $0.8176 \pm 0.0734$ & $0.8633 \pm 0.0194$ & $0.9804 \pm 0.0045$ \\
 & Proposed & 0 & $0.9421 \pm 0.0016$ & $0.9254 \pm 0.0061$ & $0.8864 \pm 0.0023$ & $0.9833 \pm 0.0002$ \\
 & baseline & 1 & $0.9361 \pm 0.0067$ & $0.8148 \pm 0.0729$ & $0.8607 \pm 0.0213$ & $0.9784 \pm 0.0050$ \\
 & Proposed & 1 & $0.9413 \pm 0.0014$ & $0.9201 \pm 0.0074$ & $0.8844 \pm 0.0020$ & $0.9821 \pm 0.0001$ \\
 & baseline & 2 & $0.9319 \pm 0.0072$ & $0.7939 \pm 0.0757$ & $0.8495 \pm 0.0233$ & $0.9784 \pm 0.0050$ \\
 & Proposed & 2 & $0.9387 \pm 0.0013$ & $0.9107 \pm 0.0077$ & $0.8788 \pm 0.0019$ & $0.9804 \pm 0.0002$ \\
\midrule
\multirow{6}{*}{JPEG}
 & baseline & 0 & $0.9363 \pm 0.0059$ & $0.8135 \pm 0.0737$ & $0.8608 \pm 0.0202$ & $0.9804 \pm 0.0033$ \\
 & Proposed & 0 & $0.9429 \pm 0.0014$ & $0.9254 \pm 0.0040$ & $0.8878 \pm 0.0021$ & $0.9829 \pm 0.0001$ \\
 & baseline & 1 & $0.9363 \pm 0.0068$ & $0.8230 \pm 0.0742$ & $0.8622 \pm 0.0210$ & $0.9804 \pm 0.0033$ \\
 & Proposed & 1 & $0.9403 \pm 0.0010$ & $0.9205 \pm 0.0079$ & $0.8827 \pm 0.0014$ & $0.9823 \pm 0.0002$ \\
 & baseline & 2 & $0.9349 \pm 0.0065$ & $0.8176 \pm 0.0783$ & $0.8587 \pm 0.0216$ & $0.9797 \pm 0.0022$ \\
 & Proposed & 2 & $0.9370 \pm 0.0011$ & $0.9152 \pm 0.0097$ & $0.8764 \pm 0.0019$ & $0.9815 \pm 0.0002$ \\
\midrule
\multirow{6}{*}{Noise}
 & baseline & 0 & $0.9373 \pm 0.0057$ & $0.8176 \pm 0.0734$ & $0.8633 \pm 0.0194$ & $0.9804 \pm 0.0045$ \\
 & Proposed & 0 & $0.9421 \pm 0.0016$ & $0.9254 \pm 0.0061$ & $0.8864 \pm 0.0023$ & $0.9833 \pm 0.0002$ \\
 & baseline & 1 & $0.9165 \pm 0.0079$ & $0.7086 \pm 0.0851$ & $0.8037 \pm 0.0305$ & $0.9717 \pm 0.0054$ \\
 & Proposed & 1 & $0.9281 \pm 0.0032$ & $0.8336 \pm 0.0328$ & $0.8496 \pm 0.0099$ & $0.9794 \pm 0.0011$ \\
 & baseline & 2 & $0.9061 \pm 0.0075$ & $0.6627 \pm 0.0906$ & $0.7727 \pm 0.0338$ & $0.9686 \pm 0.0060$ \\
 & Proposed & 2 & $0.9273 \pm 0.0131$ & $0.6676 \pm 0.0743$ & $0.7829 \pm 0.0592$ & $0.9727 \pm 0.0018$ \\
\bottomrule
\end{tabular}%
}
\end{table}

\subsection{Robustness to Image Distortions}
\label{subsec:robustness}

\subsubsection{Experimental Design}
To evaluate the sensitivity of the proposed model to common image distortions encountered in real-world fundus imaging, we conducted a robustness study on the test set under three representative distortion types, Gaussian blur, JPEG compression, and Gaussian noise. These perturbations frequently arise in ophthalmic image acquisition and transmission, where image quality may be compromised by defocus, sensor noise, or lossy compression. 
In this experiment, each distortion was applied at progressively increasing severity levels, and all models were evaluated under the 5-fold cross-validation protocol. In Table~\ref{tab:robustness}, we illustrate Accuracy, Sensitivity, F1-score and AUC to quantify the impact of distortion on classification performance and clarify the capacity distinction of resisting disturbance between the baseline and proposed model.

\subsubsection{Robustness under image quality degradation}

Robustness to image quality degradation is a critical requirement for automated glaucoma screening systems intended for real-world deployment. In practical screening scenarios, fundus images are often acquired under non-ideal conditions and may suffer from defocus blur, slight motion artifacts, sensor noise, or lossy compression during storage and transmission. These degradations can attenuate clinically relevant structures. Subsequently, the objective of this experiment is to further test whether the decision function of the proposed \textbf{Branch3KECBAM} remains stable under clinical perturbations.
Specifically, we compared the proposed model with the baseline \textbf{Patch5Model (No AM)} under three representative degradation families: Gaussian blur, JPEG compression, and Gaussian noise, each with three severity levels ($\mathrm{lvl}=0,1,2$). The results, in Table~\ref{tab:robustness}, consistently achieved higher accuracy, sensitivity, and F1-score across nearly all settings, with substantially smaller cross-validation variance, indicating that RETFound-guided attention improves both predictive performance and reliability.

\paragraph{\textbf{Gaussian Blur:}} At $\mathrm{lvl}=1$, \textbf{Branch3KECBAM} improved accuracy from $0.9361 \pm
0.0067$ to $0.9413 \pm 0.0014$, sensitivity from $0.8148 \pm 0.0729$ to $0.9201 \pm 0.0074$, and F1-score from $0.8607
\pm 0.0213$ to $0.8844 \pm 0.0020$. The $10.5\%$ sensitivity gain is clinically significant for screening applications
where missing glaucoma cases is particularly undesirable. At $\mathrm{lvl}=2$, baseline sensitivity dropped to
$0.7939 \pm 0.0757$ while the proposed model retained $0.9107 \pm 0.0077$, with F1-score of $0.8788 \pm 0.0019$ vs.
$0.8495 \pm 0.0233$.

\paragraph{\textbf{JPEG Compression:}} Compression artifacts are common in telemedicine workflows. At
$\mathrm{lvl}=2$, \textbf{Branch3KECBAM} achieved accuracy of $0.9370 \pm 0.0011$ vs. $0.9349 \pm 0.0065$, sensitivity
of $0.9152 \pm 0.0097$ vs. $0.8176 \pm 0.0783$, F1-score of $0.8764 \pm 0.0019$ vs. $0.8587 \pm 0.0216$, and AUC of
$0.9815 \pm 0.0002$ vs. $0.9797 \pm 0.0022$. The $9.8\%$ sensitivity improvement and reduced variance indicate more
  stable generalization under compression.

\paragraph{\textbf{Gaussian Noise:}} Random pixel-level corruption proved most disruptive to fine-grained retinal patterns. At $\mathrm{lvl}=1$, the proposed model improved accuracy from $0.9165 \pm 0.0079$ to $0.9281 \pm 0.0032$,
sensitivity from $0.7086 \pm 0.0851$ to $0.8336 \pm 0.0328$, F1-score from $0.8037 \pm 0.0305$ to $0.8496 \pm 0.0099$,
and AUC from $0.9717 \pm 0.0054$ to $0.9794 \pm 0.0011$. However, at $\mathrm{lvl}=2$, the advantage diminished,
suggesting extreme noise destroys both structural boundaries and textural architectures. Additional noise-aware
training or restoration modules may be required for severe noise scenarios.

Clinically, the proposed model maintains high sensitivity under blur and compression—degradations frequently
encountered in real-world imaging pipelines—while preserving AUC above $0.98$ for all such settings with reduced
cross-fold variability. This demonstrates reduced dependence on ideal image quality and greater suitability for
practical screening environments, though severe Gaussian noise remains a challenging failure mode requiring further
investigation.

\subsection{Error Analysis}

\begin{figure}
\centering
\includegraphics[width=0.95\linewidth]{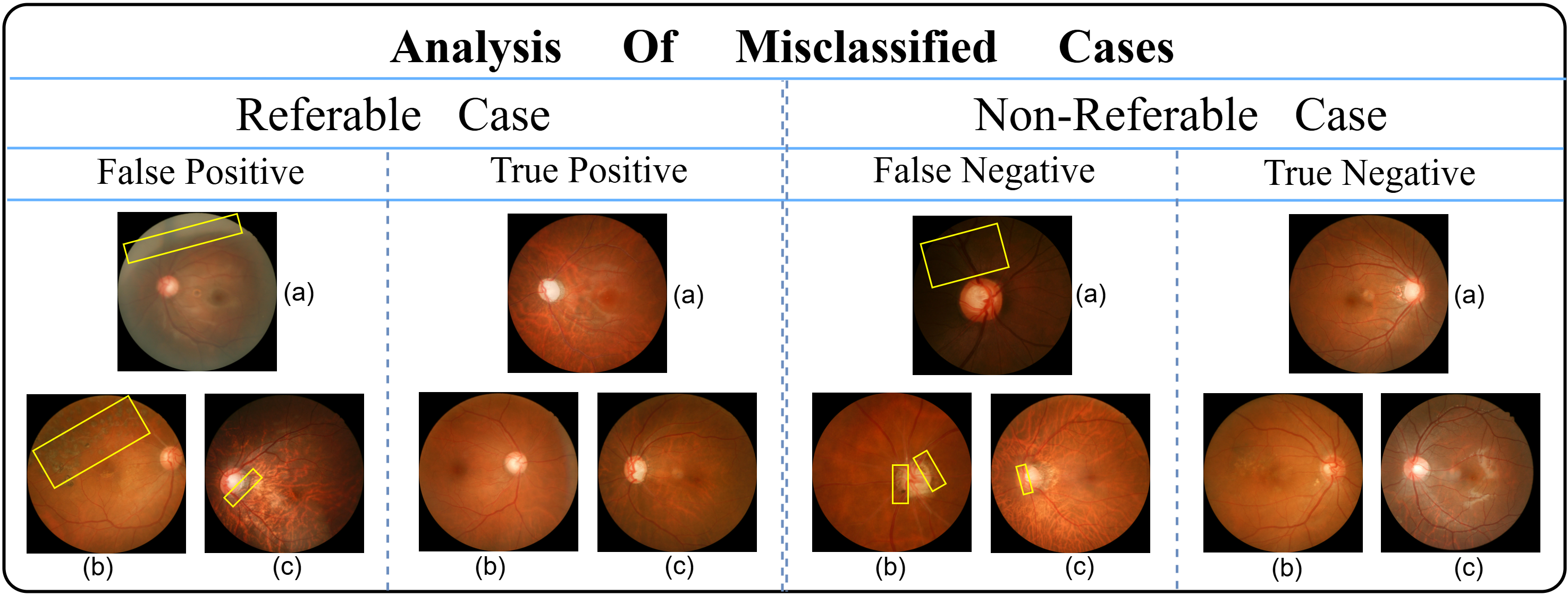}
\caption{Examples of misclassified referable and non‑referable cases. Yellow bounding boxes indicate regions that may confuse the model due to imaging artifacts or structural ambiguity.}
\label{fig5}
\end{figure}

Figure.~\ref{fig5} illustrates typical failure cases of the proposed model. Most errors originate from challenging imaging conditions rather than systematic model bias.

For referable cases, misclassification often occurs when image quality is degraded by low illumination, coarse resolution, or noise from coexisting ocular abnormalities. These factors obscure subtle pathological features and make accurate identification of disc–cup morphology difficult.

For non-referable samples, errors are primarily associated with insufficient contrast or ambiguous disc–cup boundaries. Such conditions may mimic early glaucomatous patterns and lead to false-positive predictions.

These observations indicate that the remaining errors are largely caused by visual artifacts and structural ambiguity in fundus images. Improving image quality and incorporating stronger anatomical constraints may further enhance model reliability.

\subsection{Visualization of Decision Rationale and Model Transparency}
\label{Visualization}

To further investigate the interpretability of the proposed model, we visualize the Top-k windows selected by DWM, analyze the spatial attention maps, saliency maps and feature embedding distributions.

\begin{figure}
\centering

\includegraphics[width=0.19\linewidth]{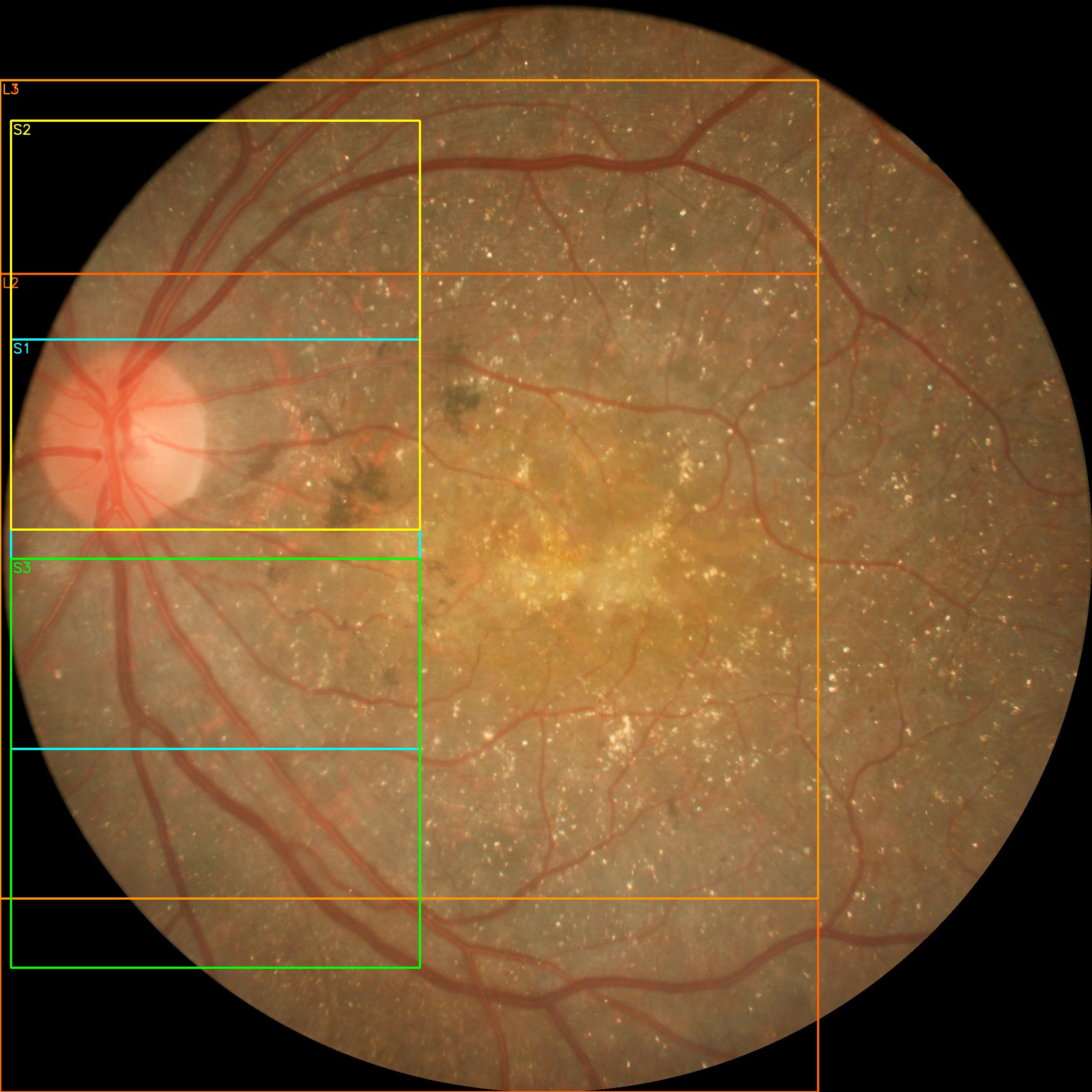}
\hfill
\includegraphics[width=0.19\linewidth]{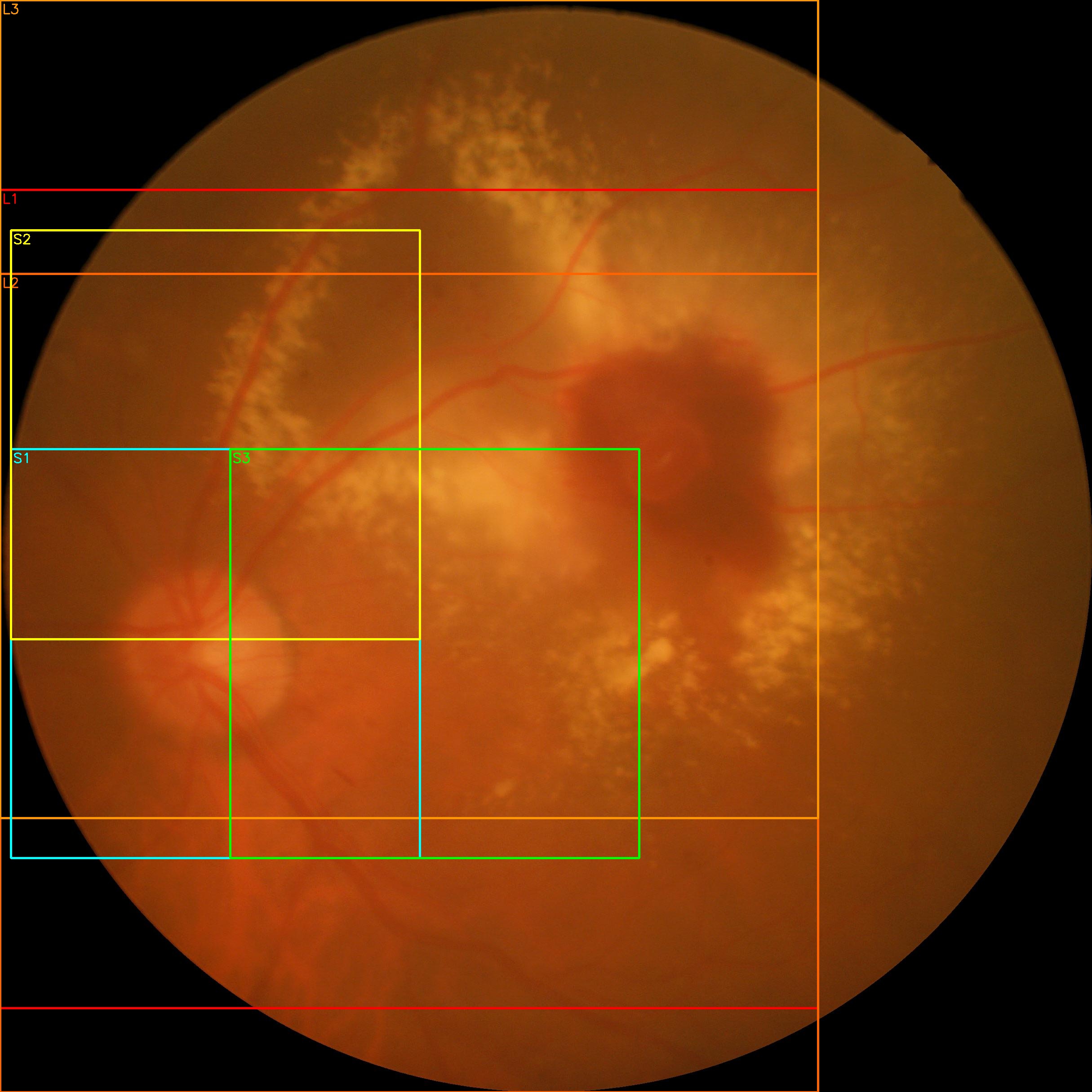}
\hfill
\includegraphics[width=0.19\linewidth]{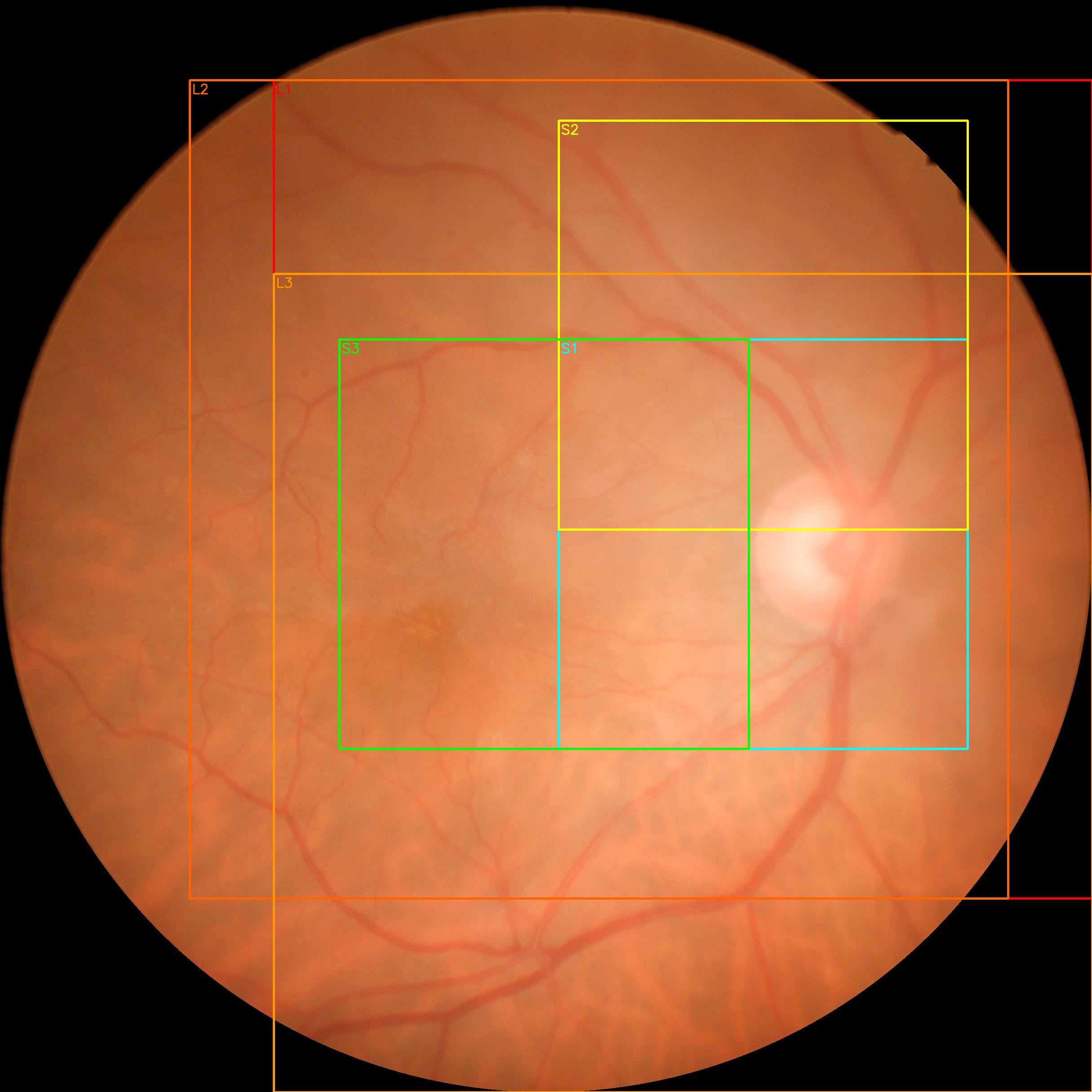}
\hfill
\includegraphics[width=0.19\linewidth]{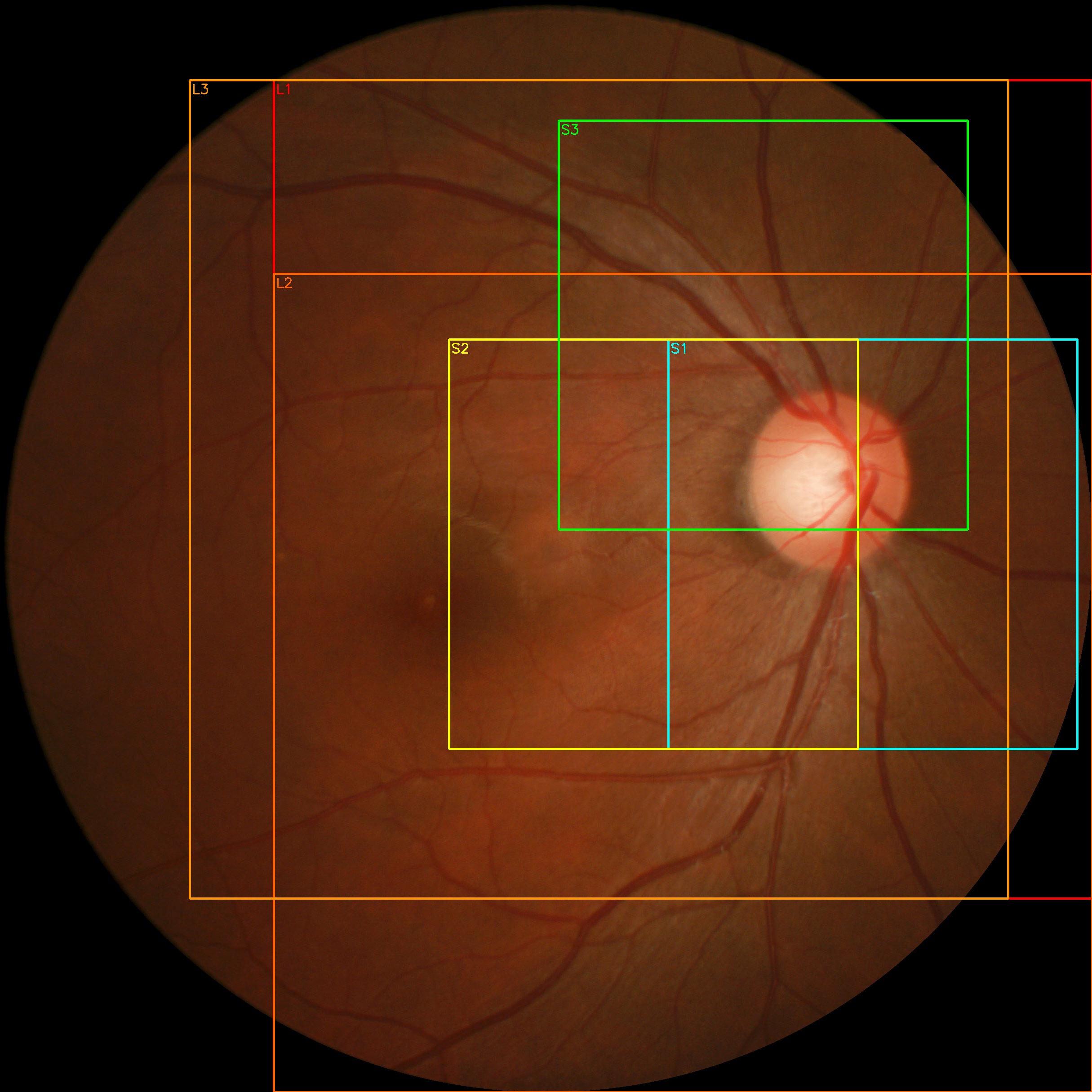}
\hfill
\includegraphics[width=0.19\linewidth]{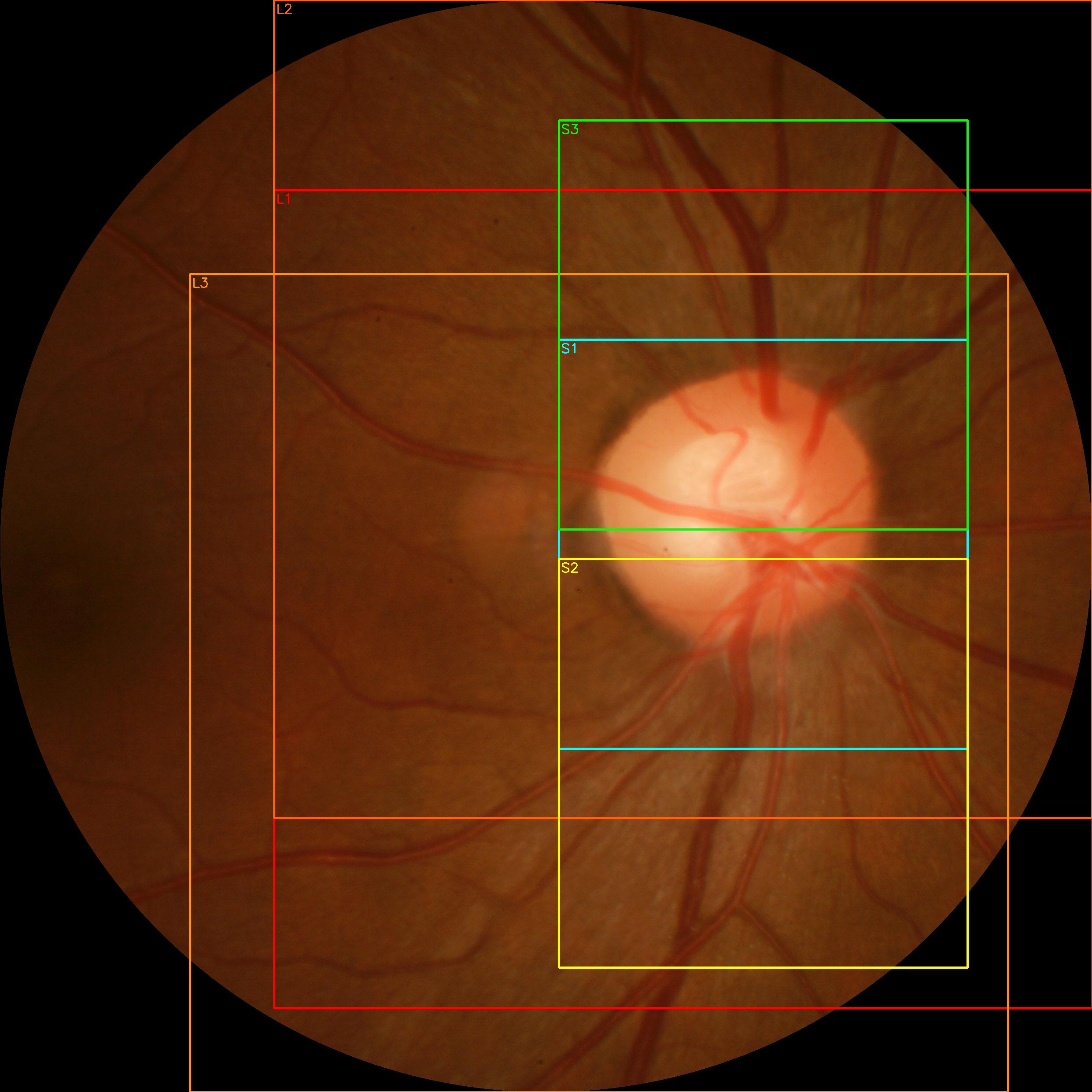}

\caption{
Top-6 windows selected by DWM on representative fundus images. Large-scale patches (\textit{L1-L3}: $224 \times 224$) and fine-grained local patches (\textit{S1-S3}: $112 \times 112$) consistently focus on the optic disc, neuroretinal rim, and major vascular regions.
}
\label{fig:dwm_visualization}
\vspace{-3mm}
\end{figure}

\paragraph{\textbf{Top-6 Windows.}} 

Figure~\ref{fig:dwm_visualization} shows the Top-6 windows selected by DWM. The selected windows consistently concentrate on clinically informative regions rather than irrelevant background areas. Large windows cover the optic disc and surrounding anatomical context, while smaller windows refine attention to vessel convergence, neuroretinal rim structures, and cup-disc boundaries, enabling DWM to capture both global structural information and local diagnostic details.

Notably, selected windows remain aligned with meaningful retinal structures even under variations in illumination, contrast,and imaging device, indicating adaptive spatial focus adjustment. Quantitative results in Table~\ref{tab:tri_binary} validatethat DWM effectively guides the model toward anatomically meaningful areas, improving both performance and interpretability.

\begin{figure}
\centering
\small

% =================== Tunable layout params ===================
\newcommand{\patchw}{0.31\linewidth}      % width of each patch image (and each header cell)
\newcommand{\patchgap}{0.02\linewidth}    % horizontal gap between patches (and header cells)

% =================== Typography helpers ===================
\newcommand{\groupheader}[1]{%
  \par\noindent{\normalsize\bfseries #1}\par\vspace{4pt}%
}

% Column headers: thinner/smaller + guaranteed spacing + aligned above each image
\newcommand{\imglabel}{%
  \par\noindent
  \fontsize{6.0pt}{7.0pt}\selectfont\bfseries
  \makebox[\patchw][c]{Orig.\strut}%
  \hspace{\patchgap}%
  \makebox[\patchw][c]{KE-CBAM\strut}%
  \hspace{\patchgap}%
  \makebox[\patchw][c]{Baseline\strut}%
  \par\vspace{2pt}%
}

% Sample caption under each case (lighter than headers)
\newcommand{\samplecap}[1]{%
  \par\vspace{1pt}\centering\footnotesize #1\par%
}

% ========================== Negative Samples ==========================
\groupheader{Non-referable Cases (Negative Samples)}

% Row 1: Neg 1 & Neg 2
\begin{minipage}{0.48\linewidth}
  \imglabel
  \centering
  \includegraphics[width=\patchw]{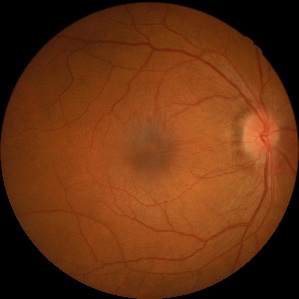}\hspace{\patchgap}%
  \includegraphics[width=\patchw]{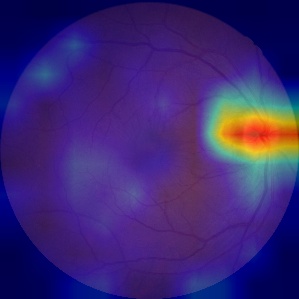}\hspace{\patchgap}%
  \includegraphics[width=\patchw]{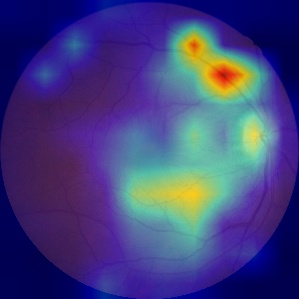}
  \samplecap{(a) Neg-Sample 1}
\end{minipage}
\hfill
\begin{minipage}{0.48\linewidth}
  \imglabel
  \centering
  \includegraphics[width=\patchw]{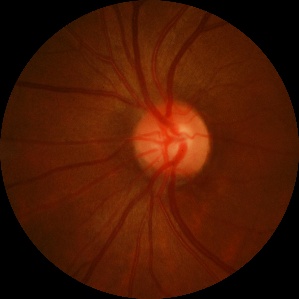}\hspace{\patchgap}%
  \includegraphics[width=\patchw]{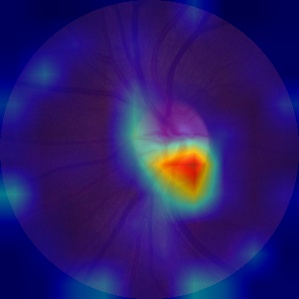}\hspace{\patchgap}%
  \includegraphics[width=\patchw]{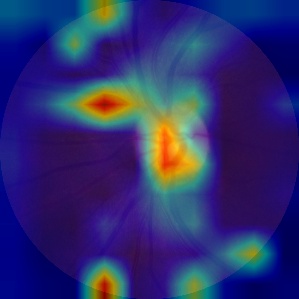}
  \samplecap{(b) Neg-Sample 2}
\end{minipage}

\vspace{5pt}

% Row 2: Neg 3 & Neg 4
\begin{minipage}{0.48\linewidth}
  \imglabel
  \centering
  \includegraphics[width=\patchw]{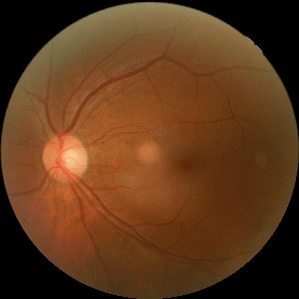}\hspace{\patchgap}%
  \includegraphics[width=\patchw]{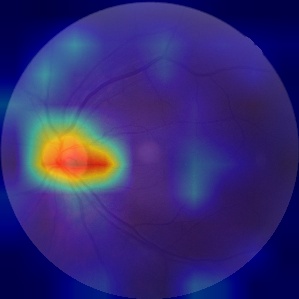}\hspace{\patchgap}%
  \includegraphics[width=\patchw]{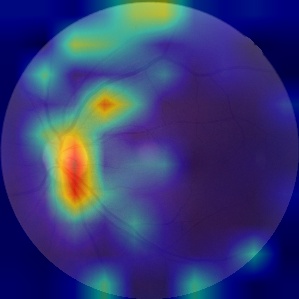}
  \samplecap{(c) Neg-Sample 3}
\end{minipage}
\hfill
\begin{minipage}{0.48\linewidth}
  \imglabel
  \centering
  \includegraphics[width=\patchw]{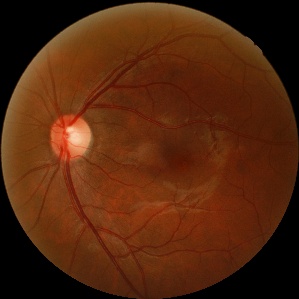}\hspace{\patchgap}%
  \includegraphics[width=\patchw]{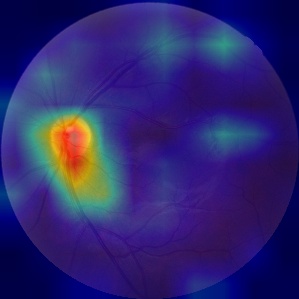}\hspace{\patchgap}%
  \includegraphics[width=\patchw]{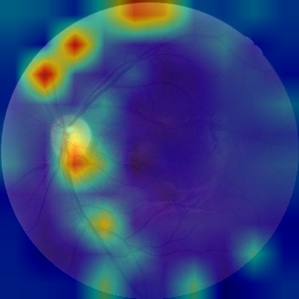}
  \samplecap{(d) Neg-Sample 4}
\end{minipage}

\vspace{8pt}

% ========================== Positive/Suspect Samples ==========================
\groupheader{Referable Cases (Positive \& Suspect Samples)}

% Row 3: Pos 1 & Pos 2
\begin{minipage}{0.48\linewidth}
  \imglabel
  \centering
  \includegraphics[width=\patchw]{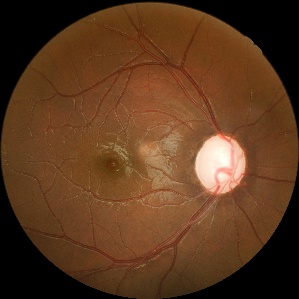}\hspace{\patchgap}%
  \includegraphics[width=\patchw]{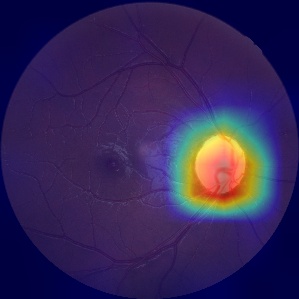}\hspace{\patchgap}%
  \includegraphics[width=\patchw]{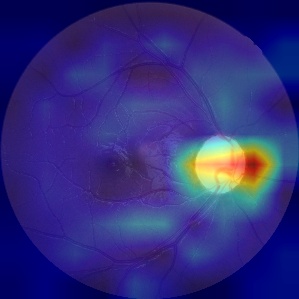}
  \samplecap{(e) Pos-Sample 1}
\end{minipage}
\hfill
\begin{minipage}{0.48\linewidth}
  \imglabel
  \centering
  \includegraphics[width=\patchw]{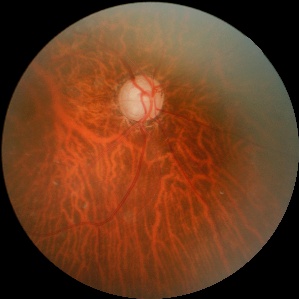}\hspace{\patchgap}%
  \includegraphics[width=\patchw]{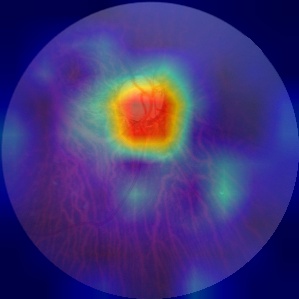}\hspace{\patchgap}%
  \includegraphics[width=\patchw]{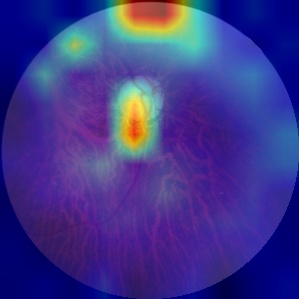}
  \samplecap{(f) Pos-Sample 2}
\end{minipage}

\vspace{5pt}

% Row 4: Pos 3 & Pos 4
\begin{minipage}{0.48\linewidth}
  \imglabel
  \centering
  \includegraphics[width=\patchw]{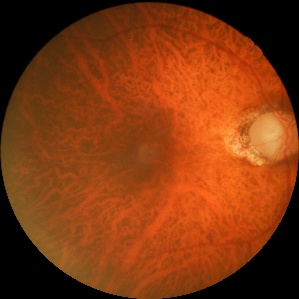}\hspace{\patchgap}%
  \includegraphics[width=\patchw]{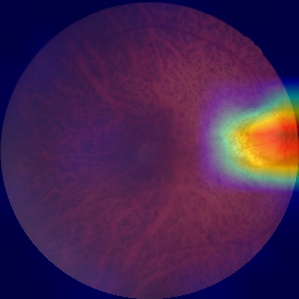}\hspace{\patchgap}%
  \includegraphics[width=\patchw]{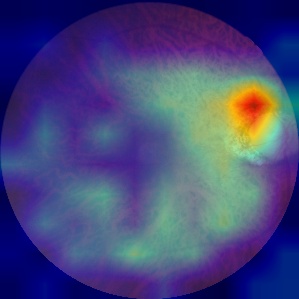}
  \samplecap{(g) Pos-Sample 3}
\end{minipage}
\hfill
\begin{minipage}{0.48\linewidth}
  \imglabel
  \centering
  \includegraphics[width=\patchw]{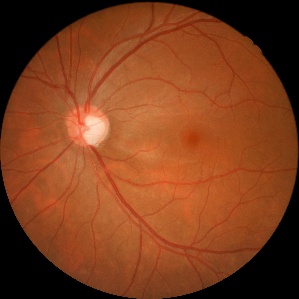}\hspace{\patchgap}%
  \includegraphics[width=\patchw]{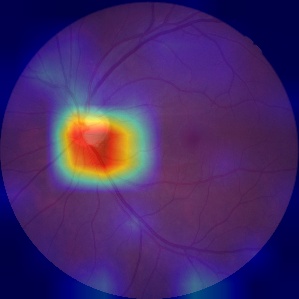}\hspace{\patchgap}%
  \includegraphics[width=\patchw]{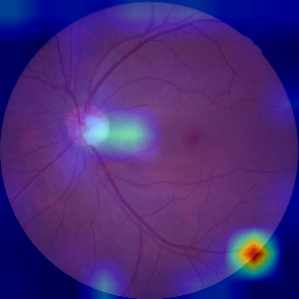}
  \samplecap{(h) Pos-Sample 4}
\end{minipage}

\caption{Saliency maps generated via Grad-CAM++ for representative glaucoma cases: Each row displays both high-quality (a, c, e, g) and low-quality (b, d, f, h) samples of each case. Columns within each group represent: Full Original, KE-CBAM CAM Map and Baseline CAM Map. The consistent focus on the optic disc and cup margins across diverse cases validates the robustness of our knowledge-enhanced attention mechanism.}
\label{fig2}
\end{figure}

\paragraph{\textbf{CAM Visualization.}} Grad-CAM++ visualizes spatial regions contributing most strongly to model predictions. As illustrated in Figure~\ref{fig2}, saliency maps are generated for both \textit{Non-referable} and \textit{Referable} categories from high-quality and low-quality fundus images, displaying original images, KE-CBAM activation maps, and CBAM activation maps.

Across all test samples, the KE-CBAM saliency maps consistently highlight the optic disc and cup boundaries, strictly aligning with clinically relevant diagnostic regions. Compared to KE-CBAM, the traditional CBAM produces more diffuse activations around the optic nerve head and mistakenly allocates higher attention weights to irrelevant background regions. This discrepancy is particularly pronounced in \textit{Referable} cases (e-h), where baseline maps exhibit widespread, unconstrained high responses across full fundus images, whereas KE-CBAM maintains concentrated attention around the OC and OD. Notably, this robust localization persists under varying image quality. For low-quality samples (b, d, f, h) suffering from low resolution, underexposure, low contrast, or imaging artifacts, the baseline model becomes highly sensitive to noise and fails to stabilize focus on optic cup regions. In contrast, KE-CBAM maintains high focal precision, directly attributable to integration of  retinal anatomical priors that explicitly guide attention toward meaningful pathological cues rather than overfitting to background textures.

To overcome the inherent subjectivity of purely qualitative visual assessments, our quantitative experiments objectively validate these interpretability outcomes. The stable focal attention observed in the Grad-CAM++ visualizations directly corroborates the objective performance improvements reported in Table~\ref{tab:tri_binary}. Specifically, the significant gains in Accuracy and AUC achieved by KE-CBAM serve as quantitative empirical evidence that the model successfully mitigates background noise and predicts relying on anatomically reliable regions.

\begin{figure}
    \centering

    % ---- A small helper: fixed-height subcaption area (1 line) ----
    \newcommand{\subcap}[1]{%
      \captionsetup{justification=centering}%
      \caption{\strut\footnotesize #1\strut}%
    }

    \begin{subfigure}[t]{0.32\linewidth}
        \centering
        \includegraphics[width=\linewidth]{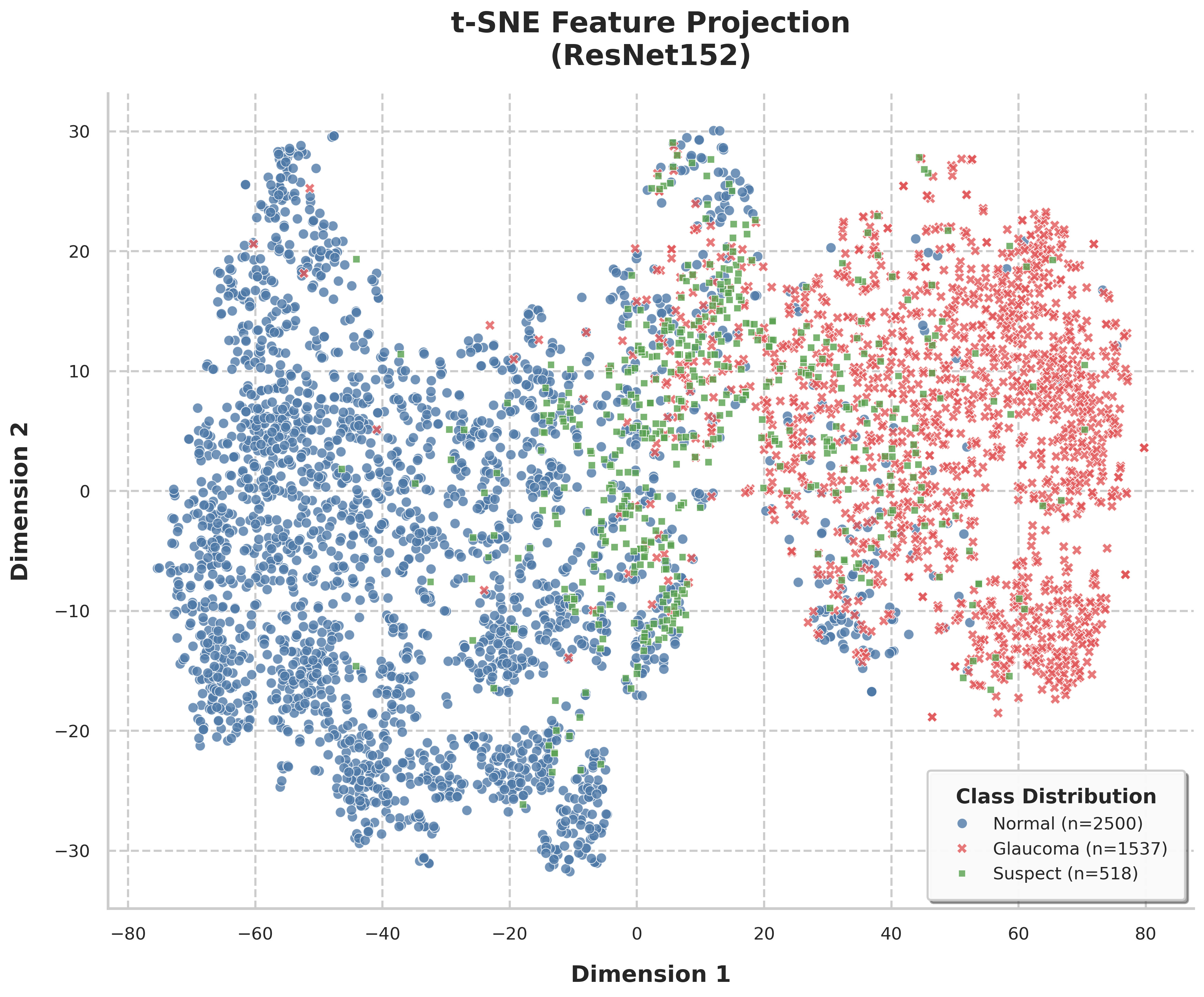}
        \subcap{ResNet152}
        \label{fig:tsne1}
    \end{subfigure}\hfill
    \begin{subfigure}[t]{0.32\linewidth}
        \centering
        \includegraphics[width=\linewidth]{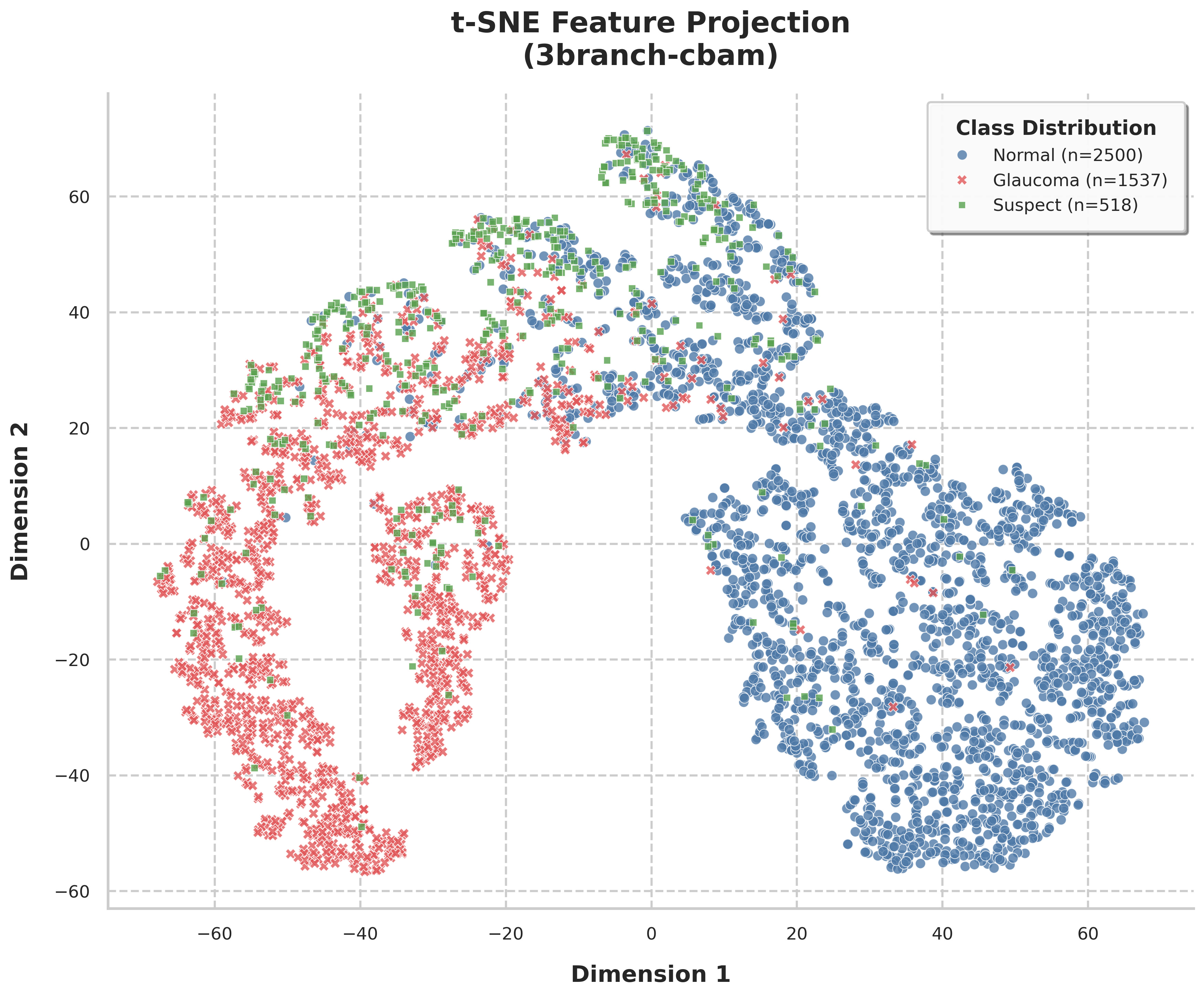}
        
        \subcap{\mbox{Branch3CBAM}}
        \label{fig:tsne2}
    \end{subfigure}\hfill
    \begin{subfigure}[t]{0.32\linewidth}
        \centering
        \includegraphics[width=\linewidth]{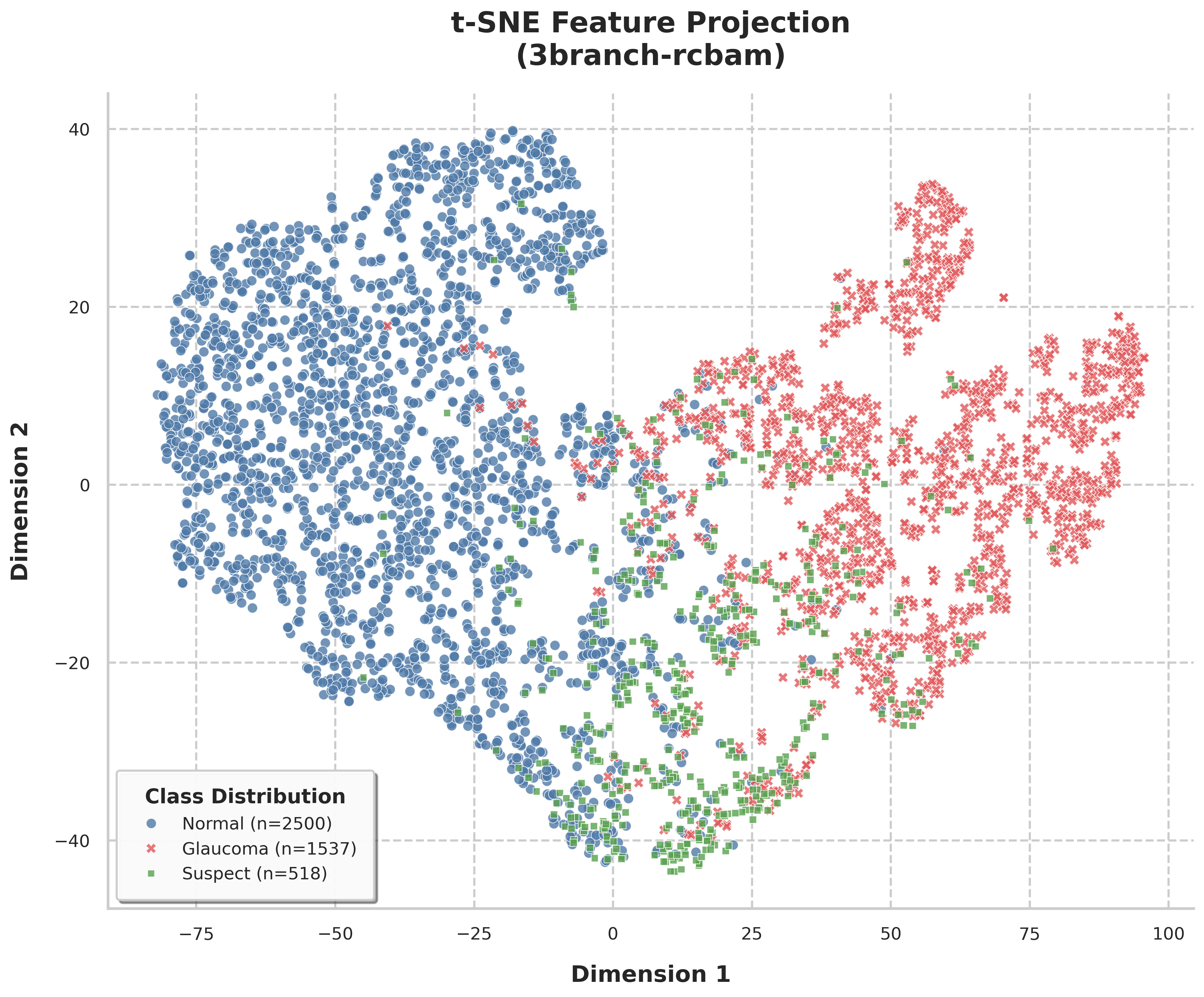}
        \subcap{\mbox{Branch3KECBAM}}
        \label{fig:tsne3}
    \end{subfigure}

    \caption{T-SNE visualization of feature embeddings learned by different model variants. The proposed KE-CBAM produces clearer inter-class separation and more compact clusters.}
    \label{fig3}
\end{figure}

\paragraph{\textbf{T-SNE Feature Embedding Analysis.}} To analyze the global structure of learned representations, we project high-dimensional embeddings into a two-dimensional space using T-SNE. As illustrated in Figure.~\ref{fig3}, the proposed KE-CBAM model produces clearer inter-class separation compared with baseline architectures. 
In the baseline model (a), feature clusters exhibit significant overlap, indicating limited discriminative capability. The introduction of CBAM partially improves cluster compactness, but inter-class mixing remains evident. In contrast, KE-CBAM forms more distinct clusters for normal and glaucomatous cases, while \textit{Suspect} samples are distributed between the two groups, reflecting the continuous nature of glaucoma progression.

These results demonstrate that knowledge-enhanced attention not only improves classification accuracy but also leads to more structured and clinically meaningful feature representations.

\section{Conclusion and Future Work}
%In this paper, we propose a tri-branch cross-attention architecture for glaucoma referability assessment, incorporating a novel KE-CBAM and a DWM. This framework comprises a global branch, a fixed ROI branch, and a DWM-based local branch. Specifically, while the global and ROI branches extract holistic semantics and anatomical context, the DWM branch adaptively identifies high-response regions potentially harboring atypical or subtle lesions. This approach effectively mitigates the uncertainty and variability inherent in optic disc–cup boundary delineation. Furthermore, the KE-CBAM integrates domain-specific pathological priors, derived from the pre-trained RETFound encoder, into the CBAM framework, enabling the network to dynamically modulate attention through the experience of medical expertise.

We propose a tri-branch cross-attention architecture for glaucoma referability assessment, incorporating a novel KE-CBAM and DWM. The framework comprises a global branch extracting holistic semantics, a fixed ROI branch capturing anatomical context, and a DWM-based local branch adaptively identifying high-response regions potentially harboring atypical or subtle lesions. This approach effectively mitigates uncertainty and variability in optic disc–cup boundary delineation. The KE-CBAM integrates domain-specific pathological priors from the pre-trained RETFound encoder into the CBAM framework, enabling the network to dynamically modulate attention through medical expertise.

Extensive evaluations across multi-source retinal datasets demonstrate that the proposed model achieves superior classification accuracy, stability, and domain generalization compared to conventional attention-based and multi-branch baselines. These results validate the efficacy of synergizing adaptive lesion localization with knowledge-driven attention for enhanced glaucoma screening. 

Future research will focus on extending this framework toward a more computationally efficient diagnostic model that adapts to real-world data and diverse clinical imaging conditions, exploring the best clinical practice of human-AI collaboration mode \cite{Wang2026CosteffectivenessHumanAI}, thereby facilitating automated glaucoma screening across heterogeneous racial and geographic populations.

\section*{Acknowledgment}
This work was supported by the National Natural Science Foundation of China under Grant No. 62402009, and the Science and Technology Development Fund of Macao under Grant No. 0013-2024-ITP1 and No. 0069/2025/ITP2.

\section*{Declaration of generative AI and AI-assisted technologies in manuscript preparation}
During the preparation of this work the authors used GPT-5.3 in order to polish writing and fix grammar errors. After using this tool/service, the authors reviewed and edited the content as needed and take full responsibility for the content of the published article.

\printcredits

\bibliographystyle{cas-model2-names}

\bibliography{cas-refs}



\end{document}